\pdfoutput=1

\documentclass[sigconf,10pt]{acmart}

\usepackage{caption}
\usepackage{setspace}
\usepackage{enumitem}

\usepackage{subcaption}
\usepackage{bbm}
\usepackage[ruled,linesnumbered,vlined]{algorithm2e}

\usepackage{tabularx}
\usepackage{multirow}

\usepackage{booktabs,subcaption,amsfonts,dcolumn}
\newcolumntype{d}[1]{D..{#1}}

\usepackage{diagbox}
\usepackage{flushend}

\usepackage{multirow}
\usepackage{multicol}

\usepackage{mathtools}

\DeclareMathOperator*{\argmax}{arg\,max}

\newlength{\myeqskip}  
\AtBeginDocument{%
    \setlength\abovedisplayskip{\myeqskip}%
    \setlength\belowdisplayskip{\myeqskip}%
    \setlength\abovedisplayshortskip{\myeqskip-\baselineskip}%
    \setlength\belowdisplayshortskip{\myeqskip}}

\acmYear{2026}\copyrightyear{2026}
\setcopyright{cc}
\setcctype[4.0]{by}
\acmConference[MobiCom '26]{The 32nd Annual International Conference on Mobile Computing and Networking}{October 26--30, 2026}{Austin, TX, USA}
\acmBooktitle{The 32nd Annual International Conference on Mobile Computing and Networking (MobiCom '26), October 26--30, 2026, Austin, TX, USA}
\acmDOI{10.1145/3795866.3844470}
\acmISBN{979-8-4007-2505-0/26/10}

\begin{document}

\title[LeanStream]{LeanStream: A Speculate-and-Refine Streaming Framework for Efficient on-Device LLM Inference}

\author{%
Renyuan Liu\textsuperscript{1},
Yuyang Leng\textsuperscript{1},
Kaiyan Liu\textsuperscript{1},
Yuzhou Zhong\textsuperscript{1},
Shaohan Hu\textsuperscript{2},
Chun-Fu (Richard) Chen\textsuperscript{2},
Peijun Zhao\textsuperscript{2},
Heechul Yun\textsuperscript{3},
Shuochao Yao\textsuperscript{1}
}

\affiliation[obeypunctuation=true]{%
  \institution{%
    \textsuperscript{1}George Mason University
    \quad
    \textsuperscript{2}Global Technology Applied Research, JPMorganChase
    \quad
    \textsuperscript{3}University of Kansas
  }%
  \city{\mbox{}}%
  \country{\mbox{}}%
}

\email{%
{rliu23,yleng2,kliu23,yzhong9}@gmu.edu,
{shaohan.hu,richard.cf.chen,peijun.zhao}@jpmchase.com
}

\email{%
heechul.yun@ku.edu,
shuochao@gmu.edu
}

\renewcommand{\shortauthors}{R. Liu et al.}

\sloppy

\begin{abstract}
On-device LLM inference improves privacy and responsiveness but remains constrained by limited device memory.
Prior systems exploit activation sparsity and offload weights to SSD or flash storage, but face a fundamental systems trade-off: accurate sparse execution decisions require the latest context, whereas efficient computation--I/O overlap requires early prediction. As a result, existing designs either serialize execution or incur redundant weight fetches, extra computation, and large cache overheads.
We present LeanStream, a streaming speculate-and-refine framework for efficient on-device LLM inference. LeanStream progressively refines computation, loading, and cache-retention priorities using partial GPU results, enabling fine-grained overlap between GPU execution and storage I/O. We implement LeanStream on both mobile and embedded platforms. Compared with prior on-device LLM inference systems, LeanStream reduces memory usage by 4.8$\times$--7.5$\times$ at the best throughput achieved by prior work, while further improving token generation throughput by 1.6$\times$--2.1$\times$. 
Source code is available at \url{https://github.com/RenyuanLIU/LeanStream}.
\end{abstract}

\keywords{Mobile Computing, On-device Inference}

\ccsdesc[500]{Computing methodologies~Machine learning}
\ccsdesc[500]{Software and its engineering}

\maketitle

{

\section{Introduction}

The recent rise of Large Language Models (LLMs) has drawn significant attention. Increasing privacy and security requirements, together with the increasing availability of everyday personal computing devices, have created strong demand for on-device LLM inference. Yet this remains challenging on mobile platforms due to their limited memory and compute capacity. To overcome this challenge, recent work has proposed storing model weights on SSDs or flash memory and dynamically activating and executing only the relevant sub-models~\cite{llmflash,powerinfer,powerinfer2}. By exploiting the widespread activation sparsity observed across a wide range of, if not all, LLMs~\cite{dejavu,liu2025trainingfree,federici2025efficient}, these approaches can dynamically identify the active weight sub-matrices, load them from storage into device memory, and compute only the necessary portions on demand.

Compared with loading and executing the full weights, exploiting dynamic activation sparsity can, in principle, reduce memory and computation costs by up to 80\%~\cite{dejavu,liu2025trainingfree,federici2025efficient}. However, realizing these theoretical gains in practice creates a fundamental tension with system-level optimizations. To make accurate decisions about which weight blocks or sub-models should be loaded and executed, the predictor ideally relies on the most recent context, namely the output of the preceding layer. Yet this dependence on fully updated context limits opportunities for optimizations such as I/O prefetching and pipelining~\cite{guo2023sti,wang2025d2moe,chen2025ktransformers, chen2026tokenflow}, leading to substantial I/O stalls.

To mitigate this bottleneck, many systems adopt layer-wise speculative I/O fetching~\cite{dejavu,llmflash,powerinfer,powerinfer2}, where activation patterns are predicted using inputs available before the current layer finishes. This enables weight transfers to overlap with the computation of the preceding layer, thereby hiding part of the I/O latency. However, because such speculation does not use the latest context, it often produces less accurate activation predictions, causing the system to load unnecessary weight sub-matrices and execute more sub-models than required. 
Some systems further mitigate I/O latency by introducing in-memory weight caches~\cite{powerinfer2,llmflash}. Yet the challenge remains similar: the most informative features for deciding which weight sub-matrices to retain and execute are often still being produced within the ongoing computation kernel. Without accurate predictive guidance for cache retention and execution prioritization, these designs can lead to substantial memory redundancy (e.g., around 3 GB for cached weights and predictive models for a 7B LLM) and significant computational overhead (e.g., more than 3$\times$ the computation required under the ideal activation pattern)~\cite{powerinfer2,llmflash}.

\begin{figure}[!t]
\vspace{-0.2cm}
        \includegraphics[width=\linewidth]{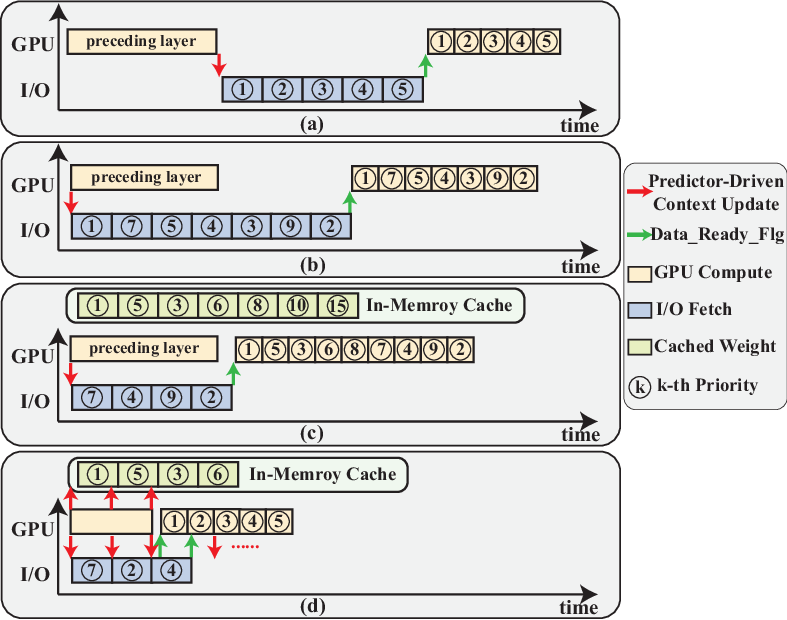}  
        \vspace{-0.7cm}
        \caption{Challenges in on-device LLM inference with activation sparsity.
        (a) Full-context prediction achieves high accuracy but forces serialized compute and I/O.
        (b) Layer-wise speculative fetching overlaps I/O with compute, but incurs prediction errors and redundant I/O.
        (c) Speculation with caching reduces I/O stalls, but still introduces memory redundancy and extra computation.
        (d) Our streaming speculate-and-refine pipeline progressively updates context for finer-grained compute–I/O coordination.
        }
        \label{fig:intro}
\end{figure}

Therefore, instead of forcing a choice between "waiting for the latest context to generate the most accurate and efficient computation and I/O decisions" and "performing long-horizon prediction to prioritize future computation and data movement", our key insight is to unify the two. We always begin with an initial prediction based on the most up-to-date information available, and then continuously refine computation prioritization and data loading/retention decisions in a fine-grained streaming manner.
Concretely, while the GPU is still processing the preceding layer, the system leverages partial intermediate results to progressively refine speculative data-loading and execution priorities, continuously updating I/O requests to reduce redundancy. At the same time, the I/O subsystem streams the required weights in fine-grained blocks, enabling the GPU to start executing the highest-priority available sub-models as soon as compute resources are free, instead of waiting for the entire sparse weight set of a layer to arrive.

However, implementing the proposed streaming framework raises several non-trivial challenges. First, the speculate-and-refine design requires frequent coordination between heterogeneous processors on mobile and embedded SoCs, with the CPU handling activation-sparsity prediction and I/O while the GPU performs computation. Existing synchronization mechanisms, such as global barriers (e.g., \texttt{cudaDeviceSynchronize} in CUDA or \texttt{clFinish} in OpenCL), incur prohibitive overhead when used at high frequency. Event-based primitives (e.g., \texttt{cudaEvent\_t} or \texttt{cl\_event}) reduce this overhead, but they provide only unidirectional notification from the co-processor to the CPU host. This limitation prevents the host from exerting real-time, bidirectional control over prioritized GPU execution based on dynamic I/O progress. In addition, because these primitives operate at kernel granularity, they often require manual kernel partitioning, which further increases kernel launch overhead and overall system cost.

More importantly, although finer-grained inter-processor communication can improve activation-pattern prediction accuracy and reduce idle time, overly frequent coordination sacrifices hardware parallelism and introduces additional overhead. The system therefore must determine an appropriate coordination frequency. Yet this choice cannot be fixed offline, because I/O latency is non-deterministic and varies with runtime factors such as cache miss behavior. As a result, execution time across the streaming pipeline becomes unpredictable, making adaptive online control essential.

The second challenge is to efficiently leverage GPU intermediate results to predict dynamic activation patterns and their relative importance. In contrast to prior methods that make predictions at kernel granularity, our approach operates at a much higher coordination frequency, significantly tightening the latency budget for prediction. If this process is not sufficiently efficient, prediction latency can itself become a bottleneck, undermining the gains from frequent inter-device coordination. This makes a lightweight, high-performance predictive model critical for system control, so that decision-making can keep pace with the high-rate speculate-and-refine stream.

To address these fundamental bottlenecks, we propose \textbf{LeanStream}, a streaming speculate-and-refine framework for efficient on-device LLM inference. LeanStream bridges the gap between context-aware prediction and system-level efficiency by enabling computation and data-movement decisions to be made progressively rather than monolithically.
LeanStream starts from an initial prediction using the freshest available context, then continuously updates computation priorities and data loading/retention decisions as intermediate results are produced. 
This design enables fine-grained overlap among I/O and GPU computation, thereby reducing redundant data movement, minimizing processor idle time, and improving end-to-end inference efficiency on resource-constrained mobile and embedded platforms.


\textit{Fine-Grained Streaming Control}. To support high-frequency information exchange, we design a lightweight communication and data-management framework that minimizes coordination overhead between GPU execution and CPU-side control and I/O. The framework reduces both synchronization overhead and metadata traffic, enabling high-grained streaming without incurring additional stalls. We also rigorously analyze the trade-off introduced by frequent coordination and design an adaptive online controller to manage it. By dynamically adjusting the synchronization frequency at runtime, the controller maintains an effective balance between prediction accuracy and hardware parallelism, thereby maximizing end-to-end system throughput.

\textit{Lightweight System Control with Stacked Learnable Hashing}. We design a lightweight control mechanism based on stacked learnable hashing for low-latency, memory-efficient prediction under fine-grained streaming execution. Compared with conventional shallow-MLP controllers, stacked learnable hashing offers high expressive capacity with substantially lower runtime and memory overhead. By relying on efficient bitwise operations, in-register table lookups, and compact output representations, it minimizes prediction latency while preserving strong modeling power. At the same time, it remains fully compatible with standard supervised learning and can be trained end-to-end with backpropagation. This makes it an effective control primitive for high-rate speculate-and-refine execution.

\begin{figure}[!t]
\vspace{-0.2cm}
        \includegraphics[width=0.75\linewidth]{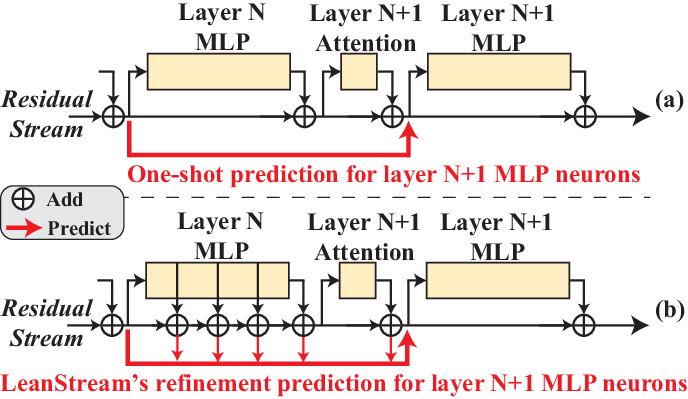}  
        \vspace{-0.4cm}
\caption{
Dependency chain of one-shot prediction and LeanStream's
partial-feature refinement.
(a) Existing layer-wise predictors use the residual state entering $\mathrm{MLP}_{n}$ to make a fixed single prediction for the neuron priorities of $\mathrm{MLP}_{n+1}$. 
(b) LeanStream executes $\mathrm{MLP}_{n}$ in priority-ordered stages. As each stage completes, its output is accumulated with the residual state to form an increasingly complete partial residual feature, which is used to refine the neuron priorities of $\mathrm{MLP}_{n+1}$. 
}
\label{fig:dependency}

\end{figure}

We evaluate LeanStream on three LLMs: Mistral-7B~\cite{mistral}, Llama2-7B~\cite{llama}, and Qwen2.5-7B~\cite{qwen}, across two embedded platforms, NVIDIA Jetson AGX Orin and Jetson AGX Xavier, and one mobile platform, the OnePlus 13. Compared with state-of-the-art LLM inference systems~\cite{powerinfer2,llmflash,dejavu}, LeanStream reduces memory usage by 4.8$\times$ to 7.5$\times$ at the best throughput achieved by prior work, and further improves token generation throughput by 1.6$\times$ to 2.1$\times$.

In summary, this paper makes the following contributions:

\begin{itemize}[leftmargin=*,nosep]
\item We present \emph{LeanStream}, a streaming speculate-and-refine framework for efficient on-device LLM inference.

\item We introduce a \emph{fine-grained streaming control} design with adaptive online coordination across CPU, GPU, and I/O to reduce stalls while preserving hardware parallelism.

\item We propose \emph{stacked learnable hashing}, a lightweight predictive mechanism for fast system control under tight latency and memory budgets.

\item We implement and evaluate LeanStream on mobile and embedded platforms, showing substantial gains in resource utilization and inference efficiency over prior approaches.

\end{itemize}

\vspace{-0.3cm}
\section{Motivation \& Related Work}
\vspace{-0.1cm}
\subsection{Challenges of on-Device LLM Inference}

Deploying LLMs on edge and mobile devices is fundamentally limited by the gap between model size and available DRAM. In practice, this constraint is even more severe because the operating system and active applications already occupy a substantial fraction of device memory. Prior work has explored various techniques to improve the efficiency and deployability of learning systems on mobile devices~\cite{yao2017deepiot,yao2018fastdeepiot,liu2024dynaspa, liu2025daf, liu2025device, leng2023scaleflow, leng2026physical}. To address the more direct challenge of model weights exceeding available DRAM, a common approach is therefore to place model weights on SSDs and fetch them on demand~\cite{llmflash,dejavu,powerinfer,powerinfer2,federici2025efficient}. 
Most modern LLMs use decoder-only Transformers, where feed-forward networks (FFNs) dominate model size. In recent Group Query Attention models~\cite{roumeliotis2023llama}, FFNs account for roughly 80\% of parameters in Llama3-8B, Qwen2-7B, and Mistral-7B. Dynamic sparsity exploits matrix-vector-dominated token generation and the many \emph{zero-valued or near-zero elements} produced by ReLU-family~\cite{dejavu,powerinfer,song2025prosparse} and SwiGLU~\cite{zhang2022moefication,federici2025efficient,powerinfer2} activations. Exploiting these \emph{sparsity patterns} at different levels~\cite{federici2025efficient} can skip around 80\% of unnecessary computation and reduce data movement with negligible accuracy loss~\cite{federici2025efficient,powerinfer2}. Moreover, \emph{activation magnitudes} indicate the \emph{relative priority} of loading and computing corresponding weight sub-matrices. Yet translating this opportunity into end-to-end system gains remains challenging.

Figure~\ref{fig:intro} illustrates the core systems tension in sparse on-device LLM inference. In Figure~\ref{fig:intro}(a), full-context prediction provides the most accurate activation decisions but serializes GPU computation and I/O, leaving hardware underutilized. Figure~\ref{fig:intro}(b) instead predicts the next layer before the current layer finishes, overlapping I/O with computation at the cost of lower prediction accuracy, redundant weight fetching, and extra computation. Adding an in-memory cache in Figure~\ref{fig:intro}(c) mitigates I/O stalls but still incurs memory and computation redundancy because decisions rely on incomplete context.

LeanStream, shown in Figure~\ref{fig:intro}(d), avoids this trade-off by progressively refining computation, loading, and cache-retention priorities as intermediate results become available. This enables fine-grained compute--I/O coordination while reducing redundant data movement and computation.

Figure~\ref{fig:dependency} illustrates partial-feature refinement. Let $\mathbf{x}_{n}$ denote the residual state before $\mathrm{MLP}_{n}$. Prior approaches use $\mathbf{x}_{n}$ for a one-shot prediction of the neuron and weight priorities of $\mathrm{MLP}_{n+1}$. LeanStream instead partitions $\mathrm{MLP}_{n}$ into $K$ stages and loads and computes them in descending predicted-priority order. Let $\mathbf{h}_{n}$ be the input to $\mathrm{MLP}_{n}$, and let $\mathbf{W}_{n,\mathrm{gate}}^{(k)}$, $\mathbf{W}_{n,\mathrm{up}}^{(k)}$, and $\mathbf{W}_{n,\mathrm{down}}^{(k)}$ denote the gate-, up-, and down-projection weight slices of stage $k$. Its partial output is
\begin{equation}
\Delta\mathbf{m}_{n}^{(k)}
=
\left[
\operatorname{SiLU}
\left(
\mathbf{h}_{n}\mathbf{W}_{n,\mathrm{gate}}^{(k)}
\right)
\odot
\left(
\mathbf{h}_{n}\mathbf{W}_{n,\mathrm{up}}^{(k)}
\right)
\right]
\mathbf{W}_{n,\mathrm{down}}^{(k)} .
\end{equation}
After the first $j$ stages, LeanStream forms the updated prediction feature as
\begin{equation}
\widetilde{\mathbf{x}}_{n}^{(j)}
=
\mathbf{x}_{n}
+
\sum_{k=1}^{j}
\Delta\mathbf{m}_{n}^{(k)} .
\end{equation}
Each $\widetilde{\mathbf{x}}_{n}^{(j)}$ refines the weight priorities of $\mathrm{MLP}_{n+1}$ before $\mathrm{MLP}_{n}$ completes.
Priority-ordered execution makes informative partial outputs available earlier, improving subsequent prediction refinement.


\begin{figure}[!t]
\vspace{-0.2cm}
        \includegraphics[width=0.8\linewidth]{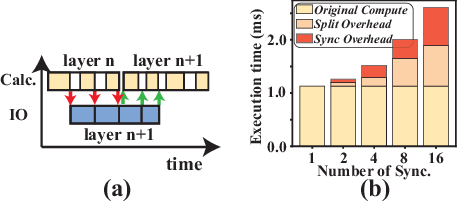}  
        \vspace{-0.4cm}
        \caption{Synchronization overhead analysis. 
        (a) Existing synchronization primitives introduce execution stalls under frequent coordination.
        (b) Increasing synchronization frequency significantly raises the GPU execution time of a Mistral-7B MLP layer with 80\% sparsity on Jetson AGX Orin.
        }
        \label{fig:motivation_sync}
\end{figure}

Realizing this design requires high-frequency coordination between GPU computation and I/O, but existing synchronization mechanisms (\texttt{cudaDeviceSynchronize} in CUDA or \texttt{clFinish} in OpenCL) are too expensive at that granularity. As shown in Figure~\ref{fig:motivation_sync}(b), increasing synchronization frequency sharply increases the execution time of a Mistral-7B MLP layer with 80\% sparsity on Jetson AGX Orin. This overhead comes from both the synchronization primitive itself and the kernel-fragmentation cost of splitting monolithic kernels into smaller schedulable units. 
These results motivate a custom low-overhead communication runtime for fine-grained speculate-and-refine execution.

\vspace{-0.2 cm}
\subsection{The Inefficiency of Static Coordination}
\vspace{-0.15 cm}
To further optimize hardware efficiency, a common design objective is to fine-tune the execution schedule to maximize the overlap between weight loading and GPU computation. This involves identifying the optimal coordination points where the system can interleave I/O requests with compute kernels without introducing significant stalls~\cite{narayanan2019pipedream,wang2022overlap,chen2024centauri,bae2021flashneuron,rajbhandari2021zero}.

However, the effectiveness of offline scheduling fundamentally depends on workload determinism, an assumption that does not hold for sparse LLM inference. As shown in Figure~\ref{fig:motivation_cachemiss}, our measurements reveal substantial runtime variability that makes static profiling ineffective. Figure~\ref{fig:motivation_cachemiss}(a) shows strong spatial heterogeneity in Mistral-7B: even when 50\% of the model weights are pinned in DRAM, cache miss rates still vary significantly across layers. Figure~\ref{fig:motivation_cachemiss}(b) further shows pronounced temporal variation within a single layer across different prompts, driven by the input-dependent nature of activation patterns.
This variability causes the timing relationship between I/O fetching and GPU computation to shift continuously at runtime. As a result, an offline schedule can quickly become suboptimal when actual cache miss behavior deviates from the profiled average, leading to either hardware underutilization or excessive I/O stalls.
These observations show that effective coordination cannot rely solely on precomputed schedules. Instead, LeanStream adopts an adaptive online strategy that dynamically adjusts coordination frequency based on real-time execution feedback.

\begin{figure}[!t]
\centering\vspace{-0.4cm}
        \includegraphics[width=0.95\linewidth]{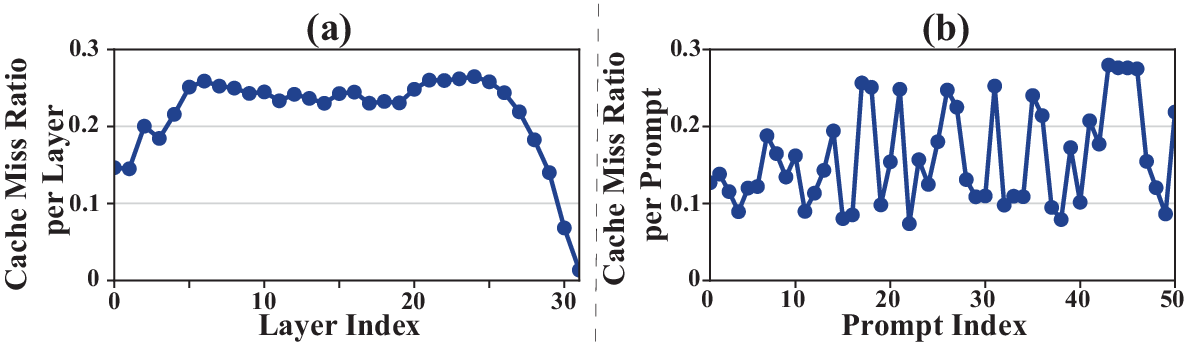}  
        \vspace{-0.3cm}
        \caption{
        Cache miss dynamics in Mistral-7B under a 50\% weight-cache budget.
        (a) Cache miss rates vary significantly across layers.
        (b) Cache miss rates for Layer 0 also fluctuate across prompts.
        This variability across layers and inputs makes offline profiling unreliable and motivates adaptive online coordination.
        }
        \label{fig:motivation_cachemiss}
\end{figure}

\vspace{-0.2cm}

\subsection{Prediction Overhead}~\label{sec:motivate_hash}
\vspace{-0.2cm}


LeanStream converts intermediate GPU results into weight-priority and cache-eviction decisions to guide loading, computation, and retention. On mobile heterogeneous platforms, this control is typically performed on the CPU to keep the GPU dedicated to execution.
However, Table~\ref{tab:prediction_overhead} shows that a state-of-the-art two-layer MLP predictor on the Jetson AGX Orin CPU can take longer than the sparse GPU MLP computation. Moreover, its cost grows linearly with finer-grained coordination.
Thus, conventional neural predictors cannot keep pace with high-frequency streaming, motivating a substantially lower-latency learnable control mechanism.

\vspace{-0.2cm}
\section{LeanStream Design}
\subsection{Overview}

Efficient on-device LLM execution requires fully utilizing GPU computation, DRAM, and SSD bandwidth under tight resource constraints. LeanStream achieves this through a fine-grained communication substrate that supports high-frequency coordination across heterogeneous system components. As illustrated in Figure~\ref{fig:overview}, LeanStream decouples computation from the rigid layer-by-layer I/O schedule used in prior sparse execution pipelines. Partial GPU results are continuously fed into the predictor to identify and reprioritize the most important weights for the next layer, allowing the storage subsystem to refine its fetch decisions on the fly. In the opposite direction, once any subset of required weights is loaded, the GPU immediately begins executing the corresponding partial computation instead of waiting for the entire layer’s weights to arrive. This bidirectional coordination overlaps computation and I/O more effectively, reducing stalls and improving end-to-end throughput.

We introduce the thread-block level fine-grained synchronization method and the streaming control strategy in Section~\ref{sec:control}. Next, we propose lightweight system control by proposing a stacked learnable hashing method in Section~\ref{sec:hash}.

\begin{table}[!t]
\centering
\footnotesize
\caption{GPU Computation vs. Prediction Overhead. CPU predictor follows the two-layer MLP design in state-of-the-art works~\cite{dejavu,powerinfer}. Latency is measured on Jetson AGX Orin for Mistral-7B with 80\% sparsity.}
\vspace{-0.3cm}
\label{tab:prediction_overhead}
\begin{tabular}{@{}cccc@{}}
\toprule
\textbf{Split} & \textbf{GPU Time} & \textbf{Pred. Time} & \textbf{Overhead} \\ 
($N$)          & ($T_{gpu}$)       & ($N \times T_{pred}$) & \textbf{Ratio} \\ \midrule
1 (Original)       & 1.13 ms           & 1.41 ms              & 1.25$\times$  \\
2              & 1.26 ms           & 2.82 ms              & 2.24$\times$  \\
4              & 1.52 ms           & 5.64 ms              & 3.71$\times$  \\
8              & 2.01 ms           & 11.28 ms             & 5.61$\times$  \\ \bottomrule
\end{tabular}
\end{table}

\vspace{-0.35cm}
\subsection{Fine-Grained Streaming Control} ~\label{sec:control}
To reduce redundancy and idle time, LLM inference should decouple GPU computation from I/O dependencies through fine-grained, context-aware coordination. Conventional approaches rely on kernel partitioning and global device synchronization, which incur substantial overhead.

We propose a thread-block level communication mechanism that is both non-blocking and asymmetric. This method leverages the unified DRAM memory architecture of SoCs to enable direct inter-device coordination without kernel splitting. In Section~\ref{sec:primitive}, we define the primitives for this thread-block level communication. Section~\ref{sec:overhead} analyzes the overhead associated with LLM execution under this communication model. Finally, Section~\ref{sec:control_strategy} presents a strategy for dynamically adjusting the communication frequency to optimize the overall LLM inference stream.

\begin{figure}[!t]
\centering\vspace{-0.2cm}
        \includegraphics[width=1\linewidth]{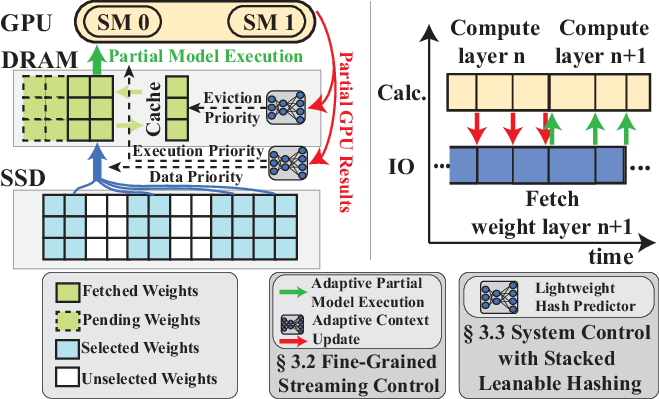}  
        \vspace{-0.6cm}
        \caption{The overview of LeanStream Framework.}
        \label{fig:overview}
\end{figure}

\subsubsection{Lightweight Coordination Primitives.}~\label{sec:primitive}

To achieve coordination that is context-aware rather than strictly bound to the execution of a specific GPU kernel, we must intervene within the kernel itself, enabling the internal scheduling of the kernel to interact with external information. Furthermore, we seek a flexible communication mechanism that avoids the mutual waiting inherent in global device-level synchronization.
This approach does not require strictly peer-to-peer or symmetric information exchange. For example, the I/O unit is not mandated to respond to every individual GPU instruction and is permitted to react to multiple GPU signals simultaneously. By allowing such decoupled interactions, the LLM inference stream becomes more flexible and the mutual interference between devices is minimized.

To address this issue, we propose a thread-block level communication mechanism that is non-blocking, asymmetric, and fine-grained. As illustrated in Figure~\ref{fig:sync}, we leverage the unified memory architecture of the SoC, which allows the GPU and CPU to access the same data simultaneously. We insert flag checks before the execution of a GPU block to verify whether the corresponding I/O block has been loaded by the CPU. Furthermore, once a block completes its computation, it updates a flag while writing its partial results back to DRAM. This design allows the GPU to utilize already loaded weights for computation while the I/O unit concurrently loads the remaining data for the current layer, ensuring that the transfer between the GPU and I/O blocks remains asynchronous as shown in Figure~\ref{fig:sync} (a).
Similarly, during I/O idle cycles, the CPU can verify completed computation results. If multiple GPU thread blocks have finished their tasks, the CPU can aggregate these results to predict and update the I/O selection. As shown in Figure~\ref{fig:sync} (b), the CPU operates asynchronously with GPU computation and can process multiple GPU results within a single update cycle.

As illustrated in Algorithm~\ref{alg:unified_logic}, our mechanism enables a fine-grained, asynchronous flow by embedding synchronization logic directly within GPU thread-blocks. In this model, each block's "leader thread" performs a non-blocking check on \texttt{gpu\_rd\_flg}. If the required weights are pre-loaded in DRAM, the block immediately initiates computation, bypassing device synchronization stalls. Upon completion, the block writes back partial results and signals its status via \texttt{gpu\_wr\_flg}, maintaining kernel persistence while providing the CPU with real-time state visibility.
The I/O worker operates in a complementary fashion by monitoring \texttt{io\_rd\_flg} for pending prediction tasks. To maintain internal consistency, a \texttt{pthread\_barrier} synchronizes worker lanes before and after parallel I/O submissions. Within each worker, \texttt{lane 0} manages global atomic flags and triggers sibling threads via a local \texttt{group\_go} variable for efficient submission. Once asynchronous I/O events complete, the worker updates \texttt{io\_wr\_flg} to release waiting GPU blocks.

\begin{figure}[!t]
\centering\vspace{-0.8cm}
        \includegraphics[width=0.95\linewidth]{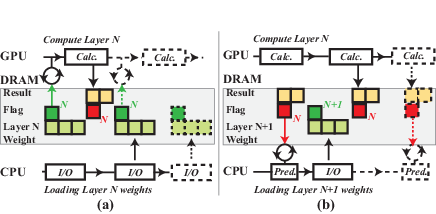}  
        \vspace{-0.3cm}
        \caption{
        Thread-block–level, non-blocking, asymmetric coordination mechanism for LLM streaming.
        (a) GPU thread-blocks perform non-blocking checks on fine-grained flags to consume weights that may have been pre-loaded into DRAM, decoupling weight arrival from compute execution. 
        (b) The CPU asynchronously monitors completed thread-blocks and aggregates multiple partial results to update I/O selection, supporting an asymmetric mapping where one CPU update responds to an arbitrary number of GPU signals.}
        \label{fig:sync}
\end{figure}

\begin{algorithm}[!t]

\small
\caption{\small Lightweight Coordination Mechanism}
\label{alg:unified_logic}
\KwIn{Atomic flags \texttt{gpu\_rd\_flg}, \texttt{gpu\_wr\_flg}, \texttt{io\_rd\_flg}, \texttt{io\_wr\_flg}, group \texttt{G}, layer \texttt{L}, iteration \texttt{N}}
\KwOut{Synchronized execution across GPU compute and libaio workers}
\BlankLine

\SetKwBlock{GPUFunc}{Function \texttt{GPU\_Kernel(layer\_id, group\_id)}}{end}
\SetKwBlock{IOFunc}{Function \texttt{I/O\_Worker(layer\_id, group\_id, lane\_id)}}{end}

\GPUFunc{
    \If{\texttt{get\_local\_id(0) == 0}}{
        \While{\texttt{atomic\_load(gpu\_rd\_flg[layer\_id][group\_id])} $\neq$ \texttt{ready}}{\textit{spin\_wait()}\;}
    }
    \texttt{barrier(CLK\_GLOBAL\_MEM\_FENCE)}\;
    \texttt{/* Execute sparse computation kernels */} \;
    \texttt{barrier(CLK\_GLOBAL\_MEM\_FENCE)}\;
    \If{\texttt{get\_local\_id(0) == 0}}{
        \texttt{atomic\_store(gpu\_wr\_flg[layer\_id][group\_id], 1)}\;
    }
}

\BlankLine

\IOFunc{
    \For{\texttt{it} $\gets 0$ \KwTo \texttt{N}}{
    \texttt{pthread\_barrier\_wait(\&iter\_start\_barrier)};
        \If{\texttt{lane\_id == 0}}{
            \While{\texttt{atomic\_load(io\_rd\_flg[layer\_id][group\_id])} $\neq$ \texttt{it}}{\textit{spin\_wait()}\;}
            \texttt{group\_go[group\_id] $\gets 1$}\;
        }
        \Else{
            \While{\texttt{group\_go[group\_id] == 0}}{\textit{spin\_wait()}\;}
        }
        
        \texttt{io\_submit(ctx, BLK\_PER\_THR, cbs)}\;

        \If{\texttt{lane\_id == 0}}{
            \texttt{atomic\_store(io\_wr\_flg[layer\_id][group\_id], 1)}\;
            \texttt{group\_go[group\_id] $\gets 0$}\;
        }
        \texttt{pthread\_barrier\_wait(\&iter\_end\_barrier)}\;
    }
}
\end{algorithm}

Crucially, the relationship between \texttt{gpu\_wr\_flg} and \texttt{io\_rd\_flg} is not necessarily a one-to-one mapping, reflecting the non-symmetric nature of our coordination substrate. To maximize throughput, the system allows for a many-to-one correspondence where multiple completed \texttt{gpu\_wr\_flg} signals can be aggregated into a single \texttt{io\_rd\_flg} update for a collective I/O prediction. Conversely, a single I/O completion signal may resolve the dependencies for multiple GPU thread-blocks simultaneously. This flexibility allows the communication frequencies of computation and I/O to differ, further decoupling the execution progress of individual hardware units.

\subsubsection{Analysis of Coordination Overheads.}~\label{sec:overhead}

While our fine-grained coordination primitive makes speculate-and-refine execution possible, prioritized loading and computation introduce additional overhead. In this section, we analyze how these mechanisms affect both I/O and computation, and use the resulting insights to motivate the coordination-scheduling design in the next section.

Prioritizing weight fetching based on real-time importance scores inherently disrupts the sequential access patterns of both storage and computation. This is because the scheduler issues I/O requests according to predicted activation magnitudes rather than the physical layout of weights on disk. As a result, out-of-order data movement can increase the number of I/O operations, reduce effective I/O block size, and degrade storage-bandwidth utilization~\cite{jeong2013stack,agrawal2008design,ji2016empirical}. It also complicates the downstream computation pipeline. To mitigate these overheads, we develop two specialized strategies.

\textit{1. I/O Placement with Co-Activated Neurons.} To mitigate the bandwidth loss caused by more frequent I/O requests and smaller transfer sizes, we optimize the physical placement of weight matrices according to neuron co-activation patterns. While prior work has noted similar effects~\cite{powerinfer2,llmflash}, existing methods largely rely on pairwise co-activation statistics and thus miss the higher-order structure required for segment-level I/O placement.

\begin{figure}[!t]
\centering\vspace{-0.2cm}
        \includegraphics[width=0.7\linewidth]{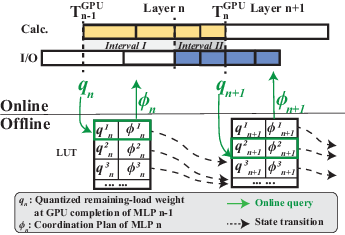}  
        \vspace{-0.3cm}
        \caption{The Streaming Control Strategy.}
        \label{fig:control}
\end{figure}

We address this by formulating I/O grouping as a distributional representation learning problem over neuron activation contexts, inspired by word embedding and contrastive representation learning~\cite{mikolov2013distributed,pennington2014glove,salakhutdinov2007learning}. Instead of using heuristic grouping rules, we learn a neuron embedding space for I/O-aware organization in sparse inference. 
The offline embedding procedure is simple: neurons that co-activate are treated as positive pairs, while neurons that do not co-activate are treated as negative pairs. For a neuron $n$ with learnable embedding $Z_n$, we sample a positive neuron $n^+$ and a negative neuron $n^-$ with learnable embeddings $Z_{n^+}$ and $Z_{n^-}$, respectively. The objective minimizes the distance between $Z_n$ and $Z_{n^+}$ while maximizing the distance to $Z_{n^-}$, as follows:
\begin{equation}
    \mathcal{L} = \min_{\{Z_n, Z_{n^+}, Z_{n^-}\}} \log(\sigma(-Z_n^T\cdot Z_{n^+})) + \log(\sigma(Z_n^T\cdot Z_{n^-}))
    \label{eqn:neuro2vec}
    \nonumber
\end{equation}
where $\sigma(\cdot)$ is a sigmoid function. The learned embeddings capture higher-order co-activation structure automatically, after which standard clustering can be used to organize neurons for I/O-aware placement. 
In practice, we group three weights into a block for Qwen2.5-7B and two weights into a block for Mistral-7B and Llama2-7B. These settings are used as the default configuration in our experiments.


\textit{2. Permutation-Invariant Execution.} On the computation side, out-of-order weight arrival would normally require complex indexing logic or dynamic kernel reconstruction~\cite{taco,dynaspa}. LeanStream avoids this overhead by exploiting the permutation invariance of the MLP hidden dimension.
For a standard SwiGLU layer, let $h_{\sigma(j)}$ be the $j$-th hidden neuron computed under permutation $\sigma$:
\[
h_{[\sigma(j)]} = \mathrm{SiLU}(X (W_1)_{[:,\sigma(j)]}) \cdot (X (W_{gate})_{[:,\sigma(j)]})
\]
The output at coordinate $m$ is the sum over these neurons:
\[
\hat{Y}_{[:,m]} = \sum_{j=1}^{k} h_{[\sigma(j)]} (W_2)_{[\sigma(j),m]} = \sum_{t=1}^{k} h_{[t]} (W_2)_{[t,m]} = Y_{[:,m]}
\]

This property allows LeanStream to treat out-of-order weights as a logically contiguous dense matrix. Because the SwiGLU output is computed as a sum over hidden neurons, it is invariant to the internal ordering of those neurons. As a result, the order in which weights arrive from storage can be used directly as the GPU execution order. The system simply appends arriving weight blocks into a contiguous memory buffer in arrival order, without any re-indexing or data reshuffling. This enables the GPU to execute high-performance dense kernels on the subset of neurons currently available.

\subsubsection{Fine-Grained Streaming Control Strategy}~\label{sec:control_strategy}

Runtime variations in sparse prediction and cache misses make a fixed
streaming schedule ineffective, while the coordination decision for the
current layer affects subsequent-layer execution. LeanStream therefore
formulates streaming coordination as a finite-horizon stochastic
predictive-control problem. Following the receding-horizon principle of
MPC~\cite{mayne2000constrained}, LeanStream optimizes from the currently
observed state and applies only the current-layer coordination plan.
Future cache and sparse-prediction behavior is represented by profiled
probability distributions and incorporated through stochastic
MPC~\cite{mesbah2016stochastic}. Solving this optimization online at
every layer boundary would be expensive. Inspired by explicit
MPC~\cite{bemporad2002explicit}, LeanStream synthesizes the
state-feedback policy offline and materializes it as a lookup table.

\textit{1. State and coordination plan.}
At $T^{GPU}_{n-1}$, let $R_n$ denote the selected weight volume of layer $n$ that remains to be loaded. The controller state is $q_n=Q(R_n)$, where $Q(\cdot)$ quantizes the remaining-load volume to a finite set of controller states.


LeanStream divides GPU computation and I/O into logical stages and
assigns each stage a communication frequency,
\begin{equation}
    \boldsymbol{\phi}_n
    =
    \left(
        \boldsymbol{\omega}_n^{G},
        \boldsymbol{\omega}_n^{\mathrm{IO}}
    \right)
    \in\Phi_n(q_n).
    \label{eq:coordination_plan}
\end{equation}
These frequencies determine the communication points that partition GPU and I/O into runtime segments.
Lower frequencies create larger segments and reduce coordination overhead. 
However, coarse GPU segments can force the GPU to wait until a larger weight chunk is loaded before proceeding, while coarse I/O segments delay the incorporation of refined GPU predictions. 
Higher frequencies allow the GPU to start computation earlier and the I/O to incorporate refined GPU predictions sooner, but incur greater coordination overhead. 
LeanStream selects $\boldsymbol{\phi}_n$ to balance these trade-offs.


\textit{2. Execution model and stochastic state transition.}
Given the controller state $q_n$ at $T^{GPU}_{n-1}$ and a candidate coordination plan $\boldsymbol{\phi}$, we derive the next state $q_{n+1}$ at $T^{GPU}_n$ from the execution between the two time points and the resulting remaining I/O workload.
We divide this execution into two intervals. Interval~I ends when the remaining layer-$n$ I/O completes. Interval~II ends when the layer-$n$ GPU computation completes. 

During Interval~I, layer-$n$ I/O does not depend on GPU progress because its final prediction has already been determined at $T^{GPU}_{n-1}$. Thus, its duration is modeled as $\bar T_n^{\mathrm I} ( q_n,\boldsymbol{\omega}_n^{\mathrm{IO}})$.
Each GPU segment can execute only after its required weight volume has been loaded. These dependencies determine the GPU work completed by the end of Interval~I, denoted as $\bar C_n^{G,\mathrm I} (\bar T_n^{\mathrm I},q_n,\boldsymbol{\phi}_n)$.
Let $C_n^G$ denote the fixed total GPU workload of layer $n$. The residual GPU workload for Interval~II is therefore $\bar C_n^{G,\mathrm{II}} = C_n^G-\bar C_n^{G,\mathrm I}$.


At the beginning of Interval~II, all weights required by layer-$n$ computation have been loaded. The GPU therefore executes the residual work without current-layer I/O dependencies, with duration $\bar T_n^{\mathrm{II}} (\bar C_n^{G,\mathrm{II}},\boldsymbol{\omega}_n^G)$, including the segmentation and communication overhead induced by the GPU coordination frequency.

Therefore, the predicted execution time from $T^{GPU}_{n-1}$ to $T^{GPU}_n$ is 
\begin{equation}
    \bar T_n(q,\boldsymbol{\phi}) = \bar T_n^{\mathrm I} + \bar T_n^{\mathrm{II}}.
\end{equation}
Given $(q_n,\boldsymbol{\phi}_n)$, the current-layer execution time is deterministic. LeanStream profiles all feasible $(q,\boldsymbol{\phi})$ configurations offline and stores the corresponding $\bar T^{\mathrm I}$, $\bar T^{\mathrm{II}}$, $\bar C^{G,\mathrm I}$, and $\bar C^{G,\mathrm{II}}$.

During Interval~II, LeanStream prefetches layer-$(n+1)$ weights until $T^{GPU}_n$.
We are interested in the remaining weight volume that still needs to be loaded at this point. 
However, even for a fixed layer $n$, residual GPU workload $\bar C_n^{G,\mathrm{II}}$, and coordination plan $\boldsymbol{\phi}_n$, different prompts can produce different predictions and cache-miss patterns. We therefore model the remaining I/O volume as a conditional distribution:
\begin{equation}
    R_{n+1}
    \sim
    \mathcal D_{
        n,\,
        \bar C_n^{G,\mathrm{II}},\,
        \boldsymbol{\phi}_n
    },
    \qquad
    q_{n+1}=Q(R_{n+1}),
    \label{eq:stochastic_transition}
\end{equation}
where $R_{n+1}$ denotes the remaining layer-$(n+1)$ weight volume at $T^{GPU}_n$ with respect to the final prediction.

In implementation, we quantize $\bar C_n^{G,\mathrm{II}}$ into 16 bins and collect 20 samples for each $(n,\bar C_n^{G,\mathrm{II}},\boldsymbol{\phi}_n)$ configuration.
Profiling all layers on Jetson AGX Orin takes approximately 90 hours, but this process is performed entirely offline and introduces no runtime decision overhead.
With the deterministic mapping from $(q_n,\boldsymbol{\phi}_n)$ to $\bar C_n^{G,\mathrm{II}}$, this conditional distribution defines the stochastic state transition from $q_n$ to $q_{n+1}$.

\textit{3. Finite-horizon stochastic optimization and explicit policy.}
The conditional distribution above induces a state-action-dependent stochastic transition. For compactness, we denote
\begin{equation}
    \mathcal D_n(q,\boldsymbol{\phi})
    \triangleq
    \mathcal D_{
        n,\,
        \bar C_n^{G,\mathrm{II}},\,
        \boldsymbol{\phi_n}
    },
    \label{eq:transition_distribution}
\end{equation}
where $\bar C_n^{G,\mathrm{II}}$ is deterministically determined by $(q,\boldsymbol{\phi})$.

LeanStream minimizes the expected latency over horizon $H$ using stochastic dynamic programming with Bellman backward recursion~\cite{marescot2013complex}:
\begin{equation}
\begin{aligned}
V_n^H(q)
=
\min_{\boldsymbol{\phi}\in\Phi_n(q)}
\Big[
    &\bar T_n(q,\boldsymbol{\phi})\\
    &+
    \mathbb E_{
        R_{n+1}\sim\mathcal D_n(q,\boldsymbol{\phi})
    }
    \left[
        V_{n+1}^{H-1}(Q(R_{n+1}))
    \right]
\Big],
\end{aligned}
\label{eq:stochastic_mpc_value}
\end{equation}
with $V_n^0(q)=0$.
The transition distributions are empirically estimated from 20 profiled samples per configuration.
The minimizing action is computed offline and stored as
\begin{equation}
    \mathrm{LUT}[n,q]
    =
    \boldsymbol{\phi}_n^{*,H}(q).
    \label{eq:stochastic_control_lut}
\end{equation}

\textit{4. Online lookup and application.}
At $T^{GPU}_{n-1}$, LeanStream measures the realized remaining-load volume and applies
\begin{equation}
    \boldsymbol{\phi}_n
    =
    \mathrm{LUT}
    \left[
        n,
        Q(R_n)
    \right].
    \label{eq:stochastic_runtime_lookup}
\end{equation}
Only the current-layer coordination plan is executed. At $T^{GPU}_n$, LeanStream observes the new realized remaining-load state and queries the corresponding policy. The prediction horizon then advances by one layer. The stochastic optimization is performed offline, while runtime control requires only state measurement and table lookup.

\vspace{-0.3cm}
\subsection{Lightweight System Control with Stacked Learnable Hashing} ~\label{sec:hash}
Our speculate-and-refine framework requires a lightweight control mechanism that can frequently use partial GPU computation results to generate control signals for data and computation prioritization. Specifically, it must estimate the relative ordering of activation magnitudes across the relevant weight sub-matrices and sub-computations in activation-sparse LLM inference, so as to guide I/O prefetching, GPU execution, and in-memory cache eviction. As discussed in Section~\ref{sec:motivate_hash}, conventional neural predictors are poorly suited for this role: they are too slow for high-frequency coordination and too memory-intensive for resource-constrained devices. Prior systems report neural predictors consuming more than 1 GB of memory~\cite{powerinfer,powerinfer2,dejavu}.
Locality-Sensitive Hashing (LSH) offers an attractive alternative because of its low computational and memory overhead. It has been successfully applied to approximate nearest-neighbor search~\cite{indyk1998approximate,gionis1999similarity,yagnik2011power}, large matrix-multiplication approximation~\cite{zeng2023lookupffn,blalock2021multiplying,wei2025t,tang2023lut}, and even large-scale neural network training~\cite{spring2017scalable,chen2020slide,chen2020mongoose}. Moreover, adaptive hash functions can further reduce LSH query cost in data-dependent or learned hashing settings~\cite{andoni2015optimal,dong2019learning,andoni2015practical}.


Our key insight is to view learnable LSH as a differentiable indexing-and-lookup-table primitive, where the adaptive hash function provides a learnable index and the table stores learnable binary features. This allows learnable LSH to replace standard neural primitives such as a single-layer MLP, but with much lower latency and memory cost. Stacking such LSH layers increases model capacity through successive nonlinear lookups while retaining efficient CPU inference, since the computation is dominated by bitwise operations and in-register table accesses~\cite{wei2025t,blalock2017bolt,wang2017survey}. In addition, when cast as a classification problem, the predictor yields compact outputs, requiring only $\log(n)$ bits for an $n$-way decision space. In contrast to conventional neural-network quantization or binarization~\cite{hubara2016binarized,xiao2023smoothquant},  our approach goes beyond reducing arithmetic precision by replacing much of the neural computation itself with learned indexing and lookup. 

\vspace{-0.3cm}
\subsubsection{Stacked Learnable Hashing}
For the $k$-th learnable hashing layer, we denote the input by $\mathbf{x}_k$ and the output by $\mathbf{y}_k$. For all layers except the first, the input $\mathbf{x}_k$ is a binary representation. Let $f_k$ be the hash function associated with layer $k$ , and let $\mathbf{T}_k \in \{\pm1\}^{2^{\tau}\times d}$ be a hash table consisting of $2^{\tau}$ buckets, each represented by a \emph{learnable} $d$-dimensional binary vector. The layer output is then defined as
\begin{equation}
    \mathbf{y}_k = \mathbf{T}_k\big[\phi_{\tau}\big( f_k(\mathbf{x_k})\big)\big] 
    \label{eqn:hash_layer}
\end{equation}
where $\phi_{\tau}(\cdot)$ converts a binary code in $\{\pm 1\}^{\tau}$ into the corresponding integer index in $\{0,1,\cdots,2^{\tau}-1\}$.

In LeanStream, we adopt hyperplane hashing~\cite{charikar2002similarity} to compute the hash code. Specifically,
\begin{equation}
    f_k(\mathbf{x_k}) = \text{sign}(\mathbf{x}_k \circledast \mathbf{W}_k)
    \label{eqn:hashfunc_layer}
\end{equation}
where $\mathbf{W}_k$ is a \emph{learnable} matrix of shape $d\times \tau$. For all layers except the first, $\mathbf{W}_k$ is also binary. Moreover, $\circledast$ denotes matrix multiplication implemented with XNOR and bitcount operations for all layers except the first.

\begin{figure*}[!t]
\vspace{-0.5cm}
        \includegraphics[width=0.9\linewidth]{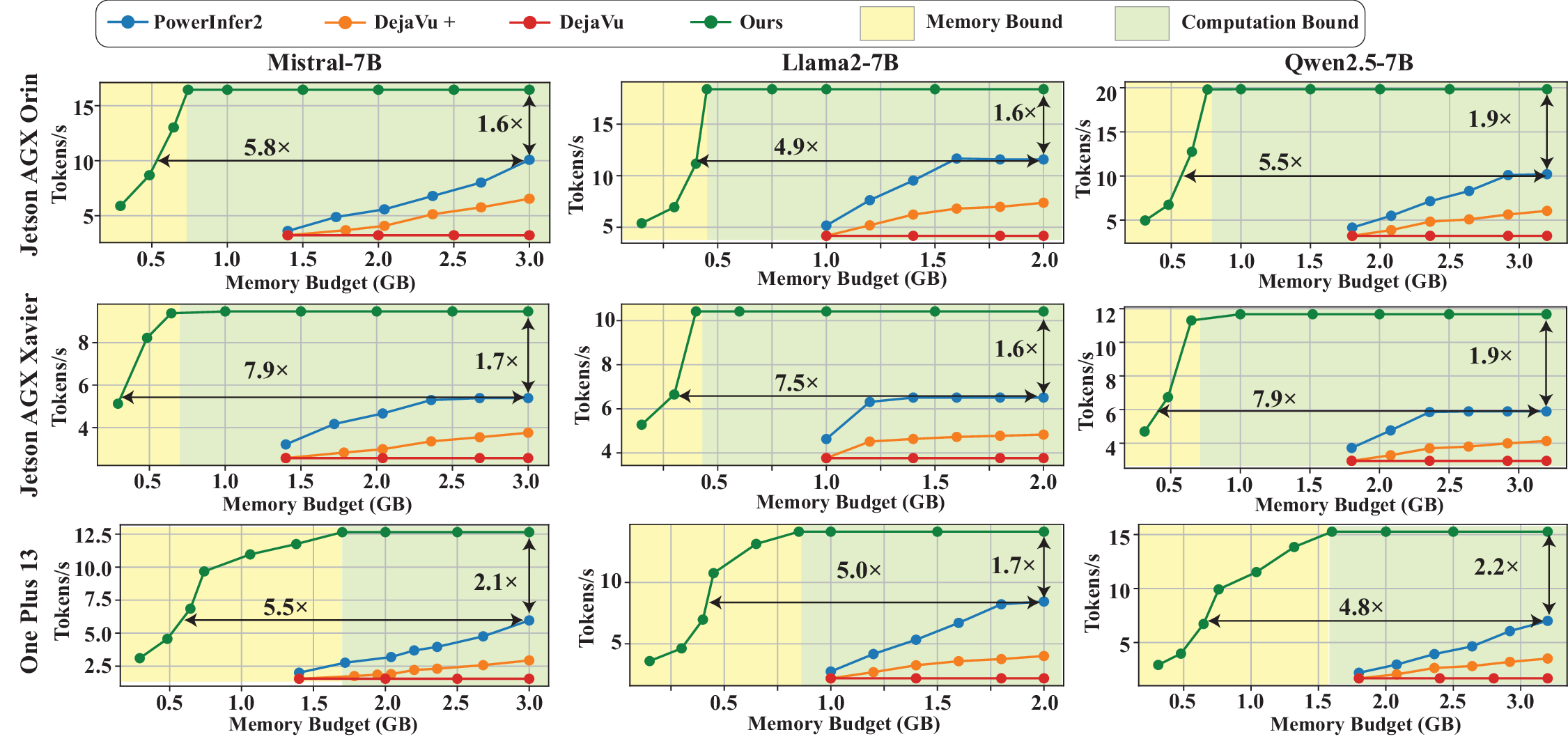}  
        \vspace{-0.5cm}
    \caption{Throughput comparison under varying memory budgets on the Scrolls-Qasper dataset across different devices and models.}
        \label{fig:end2end-all}
\end{figure*}

Stacked learnable hashing may use either a single layer or multiple layers. Unless otherwise noted, our default configuration consists of five hashing layers with $\tau=8$ and $d=256$. The final output layer is task-dependent.
For classification, the design is simple: the last learnable hashing layer outputs a hash index directly, without an additional feature table, and this index is mapped to a class label.
For regression, by contrast, the output space is quantized rather than directly cast to integers. We use equal-probability quantization~\cite{gray2002quantization}: we first profile the training labels, remove outliers via percentile-based clipping, determine the effective \texttt{min} and \texttt{max}, and estimate the corresponding cumulative distribution function (CDF). We then derive the quantization thresholds from the inverse CDF so that each quantization interval contains approximately the same probability mass.

\vspace{-0.1cm}
\subsubsection{End-to-End Differentiable Training}
To enable standard end-to-end supervised training with backpropagation, all components of stacked learnable hashing must be made differentiable. For the sign function, we adopt the standard straight-through estimator~\cite{bengio2013estimating,hubara2016binarized} during the backward pass. Another key component is the lookup operation, which is not naturally differentiable. To address this, we introduce an alternative formulation that exposes a differentiable view of the lookup and enables gradient propagation:
\begin{equation}
    \phi_{\tau}\big( f_k(\mathbf{x_k})\big) = \argmax_{i} \big(\mathbf{Z}_{[i]} \cdot f_k(\mathbf{x}_k)\big)
    \label{eqn:hashfunc_mod}
\end{equation}
where $\mathbf{Z}\in\{\pm1\}^{2^{\tau}\times\tau}$ is a structured matrix whose $i$-th row satisfies $\mathbf{Z}_{[i]} = \phi^{-1}(i)$. For example, when $\tau=3$, $\mathbf{Z}$ is an $8\times3$ matrix whose rows enumerate all possible sign patterns, i.e., $[(-1,-1,-1);(-1,-1,+1);(-1,+1,-1);\cdots;(+1,+1,+1)]$. We then apply gumbel-softmax~\cite{jang2016categorical}, a standard differentiable relaxation of $\argmax$, to enable end-to-end training. Importantly, the alternative formulation in Eq.~\eqref{eqn:hashfunc_mod} is used only during the backward pass for gradient estimation. At inference time, the compact binary representation is used directly, without expanding it into a $2^{\tau}$-dim vector.

In addition, we treat the final layer differently for classification and regression tasks. For classification, we use the same alternative formulation as in Eq.~\eqref{eqn:hashfunc_mod} and optimize it with the softmax cross-entropy loss. For regression, we find that applying binary cross-entropy loss to the quantized binary representation yields better performance, because the binary code itself preserves hierarchical similarity among quantized values.

\vspace{-0.2cm}
\subsubsection{Predictive System Control Tasks in LeanStream}
LeanStream supports two predictive tasks for system control in each LLM block. First, given a partial input feature, it predicts which neurons in the MLP output will be activated. The resulting probabilistic predictions provide a relative priority for loading the corresponding weight sub-matrices and for scheduling the computation of MLP sub-slices. Second, it predicts eviction decisions for elements in the in-memory cache. In the following, we describe the input features and output formats for these two tasks.

For neuron-activation prediction, instead of directly feeding the partial or fully updated input feature into the predictor, we first apply a PCA-based linear transformation to perform rotational dimensionality reduction while preserving the most representative directions~\cite{abdi2010principal}. The projection matrix is easily obtained from SVD, and the reduced feature representation can be computed alongside the original MLP execution by fusing this linear projection into the existing GPU kernel, resulting in negligible overhead. Similar ideas have been adopted in KV-compression systems~\cite{lee2024infinigen}. In practice, we reduce the input-feature dimension by 70\%.

For the cache-eviction task, we formulate prediction as a regression problem that estimates the reuse distance of each cache element. Combined with the element’s insertion time, the predicted reuse distance enables the system to determine its eviction priority. As input features, we use the inter-arrival times between consecutive requests to each cache object, a representation that has also been adopted in prior learning-based cache designs~\cite{song2020learning,yang2023gl}.

\vspace{-0.4cm}
\section{Evaluation}

\vspace{-0.1cm}
\subsection{Experimental Setup}
\textbf{Models and Devices.} We evaluate LeanStream on three LLMs: Mistral-7B~\cite{mistral}, Llama2-7B~\cite{llama}, and Qwen2.5-7B~\cite{qwen}. Our testbeds include two embedded platforms, NVIDIA Jetson AGX Orin and Jetson AGX Xavier, both paired with Samsung 980 Pro SSDs, as well as a mobile platform, the OnePlus 13, featuring the Snapdragon 8 Elite chipset and UFS 4.0 flash storage. This setup allows us to assess LeanStream across diverse model architectures and hardware environments.

\noindent
\textbf{Baselines.} We compare LeanStream against three primary baselines: DejaVu~\cite{dejavu}, which predicts and loads the current layer’s weights using information from the previous layer; PowerInfer-2~\cite{powerinfer2}, which combines weight prediction with in-memory weight caching to mitigate I/O latency; and DejaVu+, our enhanced DejaVu variant with an added LRU cache module to support varying memory budgets.

\noindent
\textbf{Datasets.} We conduct experiments on three representative generation tasks: long-document question answering with Scrolls-Qasper~\cite{scrolls}, open-ended factual generation with TruthfulQA~\cite{truthfulqa}, and conversational question answering with CoQA~\cite{coqa}. These tasks allow us to evaluate LeanStream across a range of context lengths and output characteristics.

\begin{figure}[!t]
\vspace{-0.3cm}
        \includegraphics[width=0.98\linewidth]{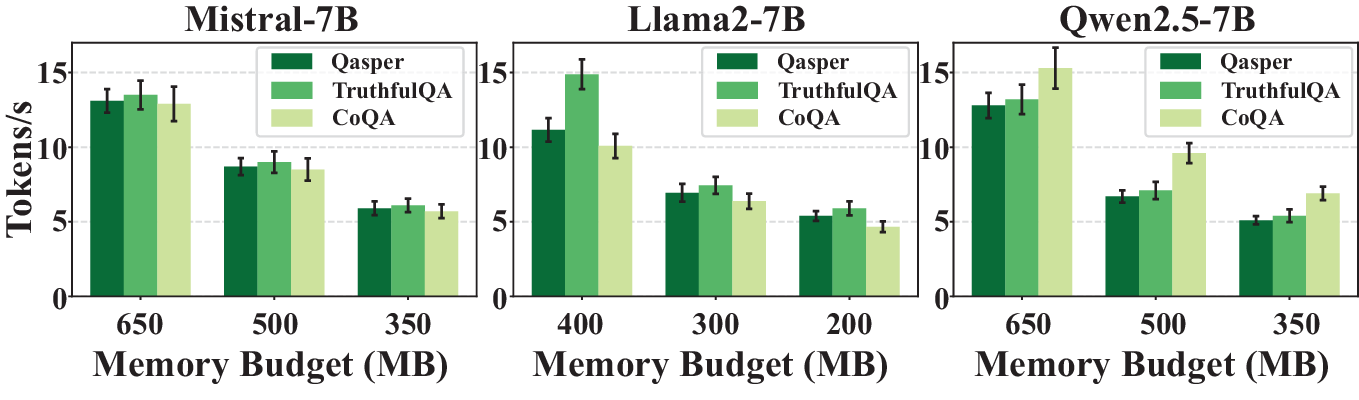}  
        \vspace{-0.3cm}
        \caption{
         Throughput comparison of our LeanStream under varying memory budgets across different datasets when streaming is memory-bound on Jetson AGX Orin.}

        \label{fig:different_dataset}
\end{figure}

\vspace{-0.5cm}
\subsection{End-to-end Results}

\textbf{Overall Performance.}
As shown in Figure~\ref{fig:end2end-all}, we compare the performance of LeanStream and the baselines on the Scrolls-Qasper dataset across different memory budgets and hardware platforms. The results indicate that LeanStream consistently outperforms all baselines in every tested scenario. Even when compared to the optimal configuration of PowerInfer-2, where 50\% of the weights are cached in memory, LeanStream achieves up to a 2.2$\times$ improvement in throughput. This performance gap becomes even more significant as the memory budget decreases. Furthermore, LeanStream can reduce the required memory budget by as much as 5.9$\times$ while still matching the peak throughput achieved by PowerInfer-2 at its optimal 50\% cache setting.

The performance characteristics vary according to the hardware capabilities of each platform. The Jetson AGX Orin exhibits a relatively balanced ratio between computation and I/O performance. The Jetson AGX Xavier features an identical SSD to the Orin, which results in excellent I/O capabilities. However, its lower computation power compared to the Orin leads to a lower peak throughput. In contrast, the OnePlus 13 utilizes UFS flash storage, which provides lower I/O bandwidth than the Jetson devices. Consequently, the OnePlus 13 enters an I/O-bound state much earlier as the memory budget decreases.

\begin{figure}[!t]
\vspace{-0.2cm}
        \includegraphics[width=0.98\linewidth]{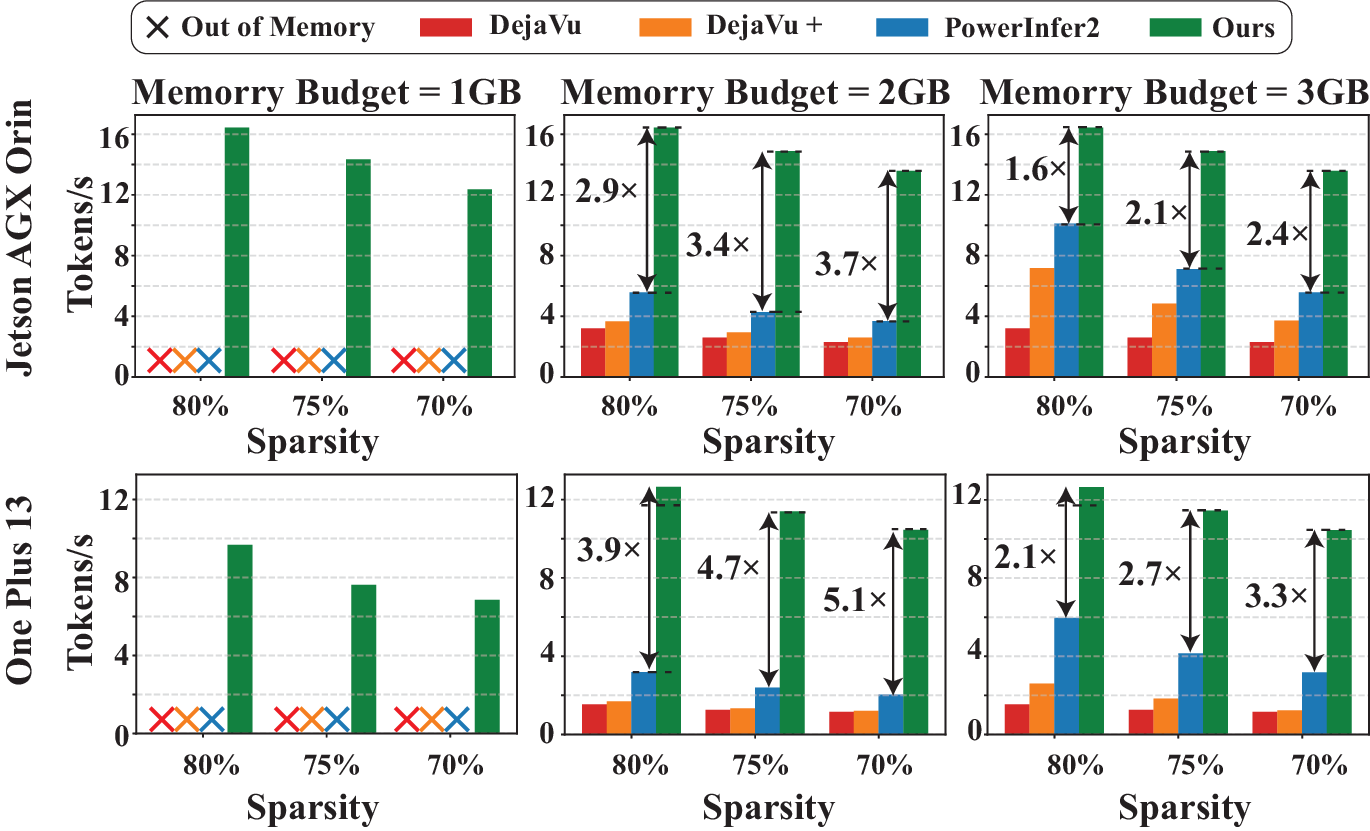}  
        \vspace{-0.3cm}
        \caption{End-to-end throughput under different sparsity levels on Qasper using Mistral-7B.}
        \label{fig:end2end_2_different_sparsity}
\end{figure}

\begin{table*}[!t]
\centering
\caption{Performance comparison under different sparsity levels.}
\vspace{-0.3cm}

\label{tab:sparsity_results}
\footnotesize
\setlength{\tabcolsep}{4pt}
\begin{tabular}{llccc|ccc|ccc}
\toprule
\multirow{2}{*}{Model} & \multirow{2}{*}{Method} 
& \multicolumn{3}{c|}{80\% Sparsity} 
& \multicolumn{3}{c|}{75\% Sparsity} 
& \multicolumn{3}{c}{70\% Sparsity} \\
\cmidrule(lr){3-5} \cmidrule(lr){6-8} \cmidrule(lr){9-11}
 &  & Qasper & TruthfulQA & CoQA 
 & Qasper & TruthfulQA & CoQA 
 & Qasper & TruthfulQA & CoQA \\
\midrule

\multirow{4}{*}{Mistral-7B}
& Original (No sparsity)     & 23.5 & 38.9 & 79.1 & 23.5 & 38.9 & 79.1 & 23.5 & 38.9 & 79.1 \\
& DejaVu       & 22.9 & 38.5 & 79.3 & 23.0 & 39.2 & 79.0 & 23.2 & 38.5 & 79.1 \\
& PowerInfer2  & 23.1 & 37.6 & 79.1 & 23.3 & 38.8 & 79.2 & 23.3 & 39.2 & 78.9 \\
& Ours         & 22.9 & 37.4 & 78.8 & 23.2 & 38.9 & 79.0 & 23.1 & 39.1 & 79.2 \\

\midrule

\multirow{4}{*}{Qwen2.5-7B}
& Original (No sparsity)      & 31.1 & 47.9 & 76.3 & 31.1 & 47.9 & 76.3 & 31.1 & 47.9 & 76.3 \\
& DejaVu       & 29.5 & 46.1 & 76.1 & 30.5 & 47.8 & 75.3 & 30.8 & 48.2 & 76.1 \\
& PowerInfer2  & 30.3 & 47.1 & 76.3 & 31.2 & 47.5 & 76.1 & 30.5 & 47.1 & 76.2 \\
& Ours         & 28.8 & 45.4 & 75.7 & 29.8 & 47.9 & 75.9 & 30.2 & 47.3 & 76.6 \\

\midrule

\multirow{4}{*}{Llama2-7B}
& Original (No sparsity)      & 25.4 & 32.2 & 77.1 & 25.4 & 32.2 & 77.1 & 25.4 & 32.2 & 77.1 \\
& DejaVu       & 25.1 & 32.1 & 77.3 & 26.1 & 31.9 & 77.8 & 25.1 & 32.4 & 77.0 \\
& PowerInfer2  & 24.8 & 31.9 & 77.1 & 25.3 & 32.3 & 76.9 & 25.3 & 31.8 & 76.9 \\
& Ours         & 24.6 & 33.2 & 76.9 & 25.1 & 33.5 & 77.0 & 24.6 & 33.7 & 76.9 \\

\bottomrule
\end{tabular}
\end{table*}


\textbf{Cross-Dataset Analysis.}
Since different datasets exhibit identical performance during compute-bound periods, Figure~\ref{fig:end2end-all} presents the overall results using Scrolls-Qasper as a representative case. Figure~\ref{fig:different_dataset} 
further compares the throughput of the three datasets on Jetson AGX Orin under memory-bound regime.
We observe that performance varies significantly depending on the combination of model and dataset even under the same memory budget. This variation occurs because the effectiveness of the prioritization mechanism in LeanStream depends on the specific activation patterns of different dataset and model combinations, which leads to different cache miss ratios. 
Consequently, these differences in cache efficiency result in distinct performance levels during the I/O-bound phase.
In this memory-bound regime, throughput is primarily dominated by cache misses, and the impact of device I/O bandwidth scales proportionally. Since all devices exhibit similar performance trends, these results on the Orin platform are representative of the behavior observed on other devices.

\begin{table}[!t]
\vspace{-0.5cm}
\centering
\caption{Synchronization overhead when partitioning one MLP layer of Mistral-7B on Jetson AGX Orin and OnePlus 13.}
\vspace{-0.3cm}
\label{tab:sync_overhead}
\footnotesize
\setlength{\tabcolsep}{2pt}
\renewcommand{\arraystretch}{0.9}

\begin{tabular}{llcccccc}
\toprule
Device & Method & \multicolumn{5}{c}{Sync. Overhead (ms)} & MLP \\
\cmidrule(lr){3-7}
 &  & 1 & 2 & 4 & 8 & 16 & Compute (ms) \\
\midrule

\multirow{2}{*}{Orin}
& Device Sync. & 0.06 & 0.13 & 0.39 & 0.88 & 1.48 & \multirow{2}{*}{1.13} \\
& Ours         & 0.01 & 0.01 & 0.12 & 0.31 & 0.47 &  \\

\midrule

\multirow{2}{*}{OnePlus 13}
& Device Sync. & 0.08 & 0.19 & 0.42 & 0.91 & 1.85 & \multirow{2}{*}{1.51} \\
& Ours         & 0.03 & 0.03 & 0.11 & 0.25 & 0.53 &  \\

\bottomrule
\end{tabular}
\end{table}

\textbf{Sparsity and Accuracy Analysis.} Table~\ref{tab:sparsity_results} presents the accuracy achieved by LeanStream and the baselines across various target weight sparsity levels for each dataset. The results demonstrate that our approach maintains accuracy levels that are nearly identical to both the original dense models and the various baselines. Figure~\ref{fig:end2end_2_different_sparsity} further illustrates the performance of various baselines under different memory budgets and sparsity levels for Mistral-7B on the Scrolls-Qasper dataset. As sparsity decreases, the overall throughput of all models declines because lower sparsity necessitates more computation and increases the volume of weights that must be loaded. Despite this, LeanStream maintains a performance lead across all tested sparsity levels. Notably, when the memory budget becomes extremely small, LeanStream remains functional while the other baselines fail due to out-of-memory errors.

\vspace{-0.4cm}
\subsection{Component Impact Study}
\vspace{-0.1cm}
\subsubsection{Analysis of Synchronization Overhead}

Table~\ref{tab:sync_overhead} compares the synchronization overhead of our proposed method against traditional device synchronization mechanisms. We evaluate these overheads when partitioning a single MLP layer of Mistral-7B on both the Jetson AGX Orin and the OnePlus 13. The results indicate that the overhead of our method is consistently lower than that of standard device synchronization. As the number of splits increases, our approach does experience an increase in overhead due to reduced parallelism. However, even when the number of splits reaches 16, our overhead remains less than half of the original computation time. In contrast, the overhead associated with traditional synchronization methods exceeds the total computation time at that same split level.

\vspace{-0.1cm}
\subsubsection{Ablation Study of Prioritization Components}

\begin{figure}[!t]
\vspace{-0.2cm}
        \includegraphics[width=0.95\linewidth]{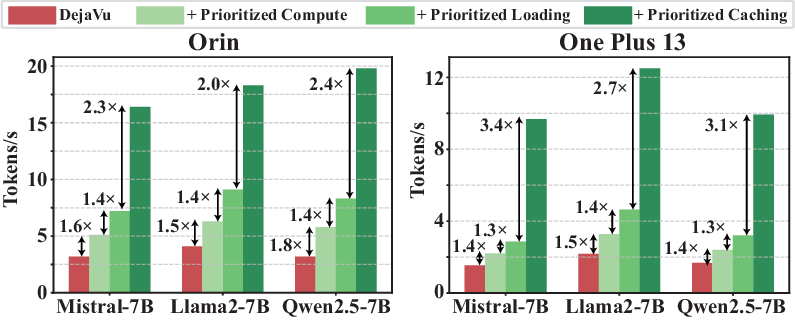}  
        \vspace{-0.2cm}
\caption{Ablation study of prioritization components on Qasper dataset with 20\% weight caching budget on Jetson AGX Orin and OnePlus 13.}
        \label{fig:prioritization}
\end{figure}

Figure~\ref{fig:prioritization} presents an ablation study of the prioritization components on the Qasper dataset, using a cache budget equal to 20\% of the model weight size, on Jetson AGX Orin and OnePlus 13. Starting from the standard DejaVu implementation, we progressively add prioritized computation, prioritized loading, and prioritized caching. This step-by-step integration allows us to isolate how each component reduces redundant computation and I/O. The results show that prioritized caching provides the largest performance gain, with an especially pronounced effect on the mobile platform where I/O bandwidth is more limited. This improvement is consistent with the low compute-to-memory-access ratio of the LLM decoding phase, in which performance is bottlenecked more by weight fetching than by arithmetic throughput. As a result, intelligently prioritizing which weights remain in cache substantially reduces high-latency I/O requests and yields the largest throughput gains.

\begin{table}[t]
\centering
\caption{Cache miss ratio of different memory management schemes for Mistral-7B on Qasper. Memory budget denotes the fraction of model weights that can be held in memory relative to the total model weights.}
\vspace{-0.2cm}
\label{tab:cache_miss_ratio}
\footnotesize
\setlength{\tabcolsep}{4pt}
\renewcommand{\arraystretch}{0.95}
\begin{tabular}{ccccc}
\toprule
Memory Budget & PowerInfer2 & LFU & LRU & Ours \\
\midrule
25\% & 0.56 & 0.76 & 0.81 & \textbf{0.11} \\
50\% & 0.19 & 0.35 & 0.39 & \textbf{0.05} \\
\bottomrule
\end{tabular}
\end{table}

\begin{table}[!t]
\centering
\footnotesize
\caption{Throughput comparison on the Qasper workload with 20\% weight caching on Jetson AGX Orin. 
\textit{Fix} uses a default fixed streaming configuration, \textit{Best Static} uses an offline-searched globally optimal static configuration of compute and I/O blocks, and \textit{Ours} applies dynamic streaming control.}
\label{tab:stream_control}
\vspace{-0.3cm}
\begin{tabular}{lccc}
\toprule
 & \multicolumn{3}{c}{Throughput (tokens/s)} \\
\cmidrule(lr){2-4}
 & Mistral 7B & Llama2-7B & Qwen2.5-7B \\
\midrule
One Shot & 6.2 & 8.9 & 10.2 \\
Best Static & 10.7 & 11.3 & 12.9 \\
Ours & \textbf{16.4} & \textbf{18.3} & \textbf{19.8} \\
\bottomrule
\end{tabular}
\end{table}

\vspace{-0.3cm}
\subsubsection{Effectiveness of Cache Policies.}
Table~\ref{tab:cache_miss_ratio} presents the cache miss ratios for various memory management schemes using Mistral-7B on the Qasper dataset across two memory budget configurations. We compare LeanStream against PowerInfer-2 and standard replacement policies including Least Frequently Used (LFU) and Least Recently Used (LRU). The results show that LeanStream achieves a significantly lower cache miss ratio than all other methods. At a 25\% memory budget, LeanStream maintains a miss ratio of only 0.11, while the baseline PowerInfer-2 and standard policies such as LRU exhibit much higher miss ratios of 0.56 and 0.81, respectively.
As the memory budget increases to 50\%, LeanStream further reduces the cache miss ratio to 0.05, which represents a nearly fourfold improvement over the PowerInfer-2 baseline. These results indicate that our prioritization mechanism is highly effective at identifying and retaining the most critical weights for inference. By minimizing cache misses, LeanStream significantly reduces the volume that must be loaded, which directly translates to the higher throughput observed in our end-to-end evaluations. 

\vspace{-0.1cm}
\subsubsection{Effectiveness of Dynamic Streaming Control}











\vspace{-0.2cm}
\begin{table}[!t]
\centering
\footnotesize
\setlength{\tabcolsep}{3pt}
\renewcommand{\arraystretch}{0.95}
\caption{
Comparison of model size, latency, and relative loading redundancy
among DNN-based, BNN-based, and our Stacked Learnable Hashing
predictors on Jetson AGX Orin.
DNN$_\mathrm{F}$ uses the full current context, while
DNN$_\mathrm{S}$ uses stale context.
}
\vspace{-0.3cm}
\label{tab:stale_vs_hash}

\begin{tabular}{c c c c c}
\toprule
& & Mistral & Llama2 & Qwen2.5 \\
\midrule

\multirow{3}{*}{Size}
& DNN  & 1.4 GB & 1.1 GB & 1.8 GB \\
& BNN  & 120 MB & 106 MB & 130 MB \\
& Ours & \textbf{24 MB} & \textbf{23 MB} & \textbf{23 MB} \\

\midrule

\multirow{3}{*}{Latency}
& DNN
& $1.41 \pm 0.03~\mathrm{ms}$
& $1.18 \pm 0.03~\mathrm{ms}$
& $1.49 \pm 0.04~\mathrm{ms}$ \\

& BNN
& $363 \pm 8~\mu\mathrm{s}$
& $325 \pm 7~\mu\mathrm{s}$
& $371 \pm 8~\mu\mathrm{s}$ \\

& Ours
& \textbf{$92 \pm 3~\mu\mathrm{s}$}
& \textbf{$87 \pm 2~\mu\mathrm{s}$}
& \textbf{$88 \pm 2~\mu\mathrm{s}$} \\

\midrule

\multirow{4}{*}{\shortstack{Relative \\ Loading \\ Redundancy}}
& DNN$_\mathrm{F}$ & 0    & 0    & 0    \\
& DNN$_\mathrm{S}$ & 35\% & 26\% & 28\% \\
& BNN              & 9\%  & 12\% & 11\% \\
& Ours             & 12\% & 14\% & 14\% \\

\bottomrule
\end{tabular}
\end{table}

Table~\ref{tab:stream_control} compares the throughput on the Qasper workload with 20\% weight caching on the Jetson AGX Orin to evaluate our streaming control mechanism. 
The "One Shot" configuration represents a baseline approach that predicts the weight only once at the input of layer.
The "Best Static" configuration uses an offline profile to select a static generally optimized granularity for I/O and computation blocks within the stream. The results demonstrate that our dynamic approach consistently outperforms the offline optimal setting. 

\vspace{-0.2cm}
\subsubsection{Efficiency of Stacked Learnable Hashing}
Table~\ref{tab:stale_vs_hash} evaluates our stacked learnable hashing
mechanism against traditional DNN-based and BNN-based prediction
methods across three key dimensions on the Jetson AGX Orin. The
results demonstrate that our hashing approach provides a superior
balance of efficiency and performance. In terms of
memory footprint, our method achieves a remarkable reduction,
requiring only approximately 23~MB to 24~MB across all tested models.
This represents a significant improvement over the 1.1~GB to 1.8~GB
required by DNN predictors and even the 106~MB to 130~MB required by
BNN implementations.

Furthermore, our approach exhibits the lowest inference latency,
processing predictions in under 100~$\mu$s. 
While BNNs show slightly lower relative loading redundancy in some cases, LeanStream maintains a highly competitive redundancy level between 12\% and 14\%. 
By drastically lowering both size and latency while maintaining effective weight selection, our stacked learnable hashing ensures a highly efficient prediction pipeline that minimizes resource contention on constrained platforms.

\vspace{-0.1cm}
\subsubsection{Validation across Refinement Steps}
\label{sec:refinement_validation}

To evaluate progressive refinement, we partition each sparse MLP into eight priority-ordered stages and update the next-layer prediction after each stage. 
We report Importance Ratio, the normalized importance captured by the predicted sparse set, and Top-10\% Recall.
Figure~\ref{fig:refinement_quality}(a) shows that both metrics improve
with progressive refinement. Compared with an otherwise identical non-priority execution order, Priority ordering provides the largest early-stage gain because high-importance neurons are computed first, while the gap narrows as both orders approach the complete sparse MLP output. Figure~\ref{fig:refinement_quality}(b) shows the same trend across five representative layers.

\vspace{-0.2cm}
\subsubsection{Analysis of Impact of Prefill Phase}

\begin{figure}[!t]
\centering
\includegraphics[width=0.75\columnwidth]{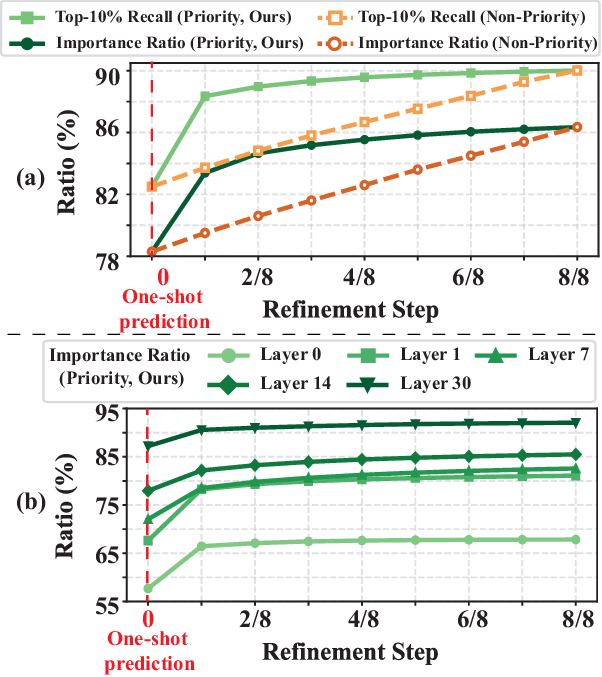}
\vspace{-0.2cm}
\caption{
Progressive prediction refinement for Mistral-7B on CoQA under 80\% activation sparsity.
LeanStream divides the MLP computation into eight stages and executes them in descending priority order. 
(a) Top-10\% recall and importance ratio averaged across all transformer layers.
(b) Per-layer importance ratio for different layers.
0/8 represents the initial one-shot prediction, while 8/8 uses the complete MLP output.
}
\label{fig:refinement_quality}
\end{figure}


\begin{figure}[!t]
\centering
        \includegraphics[width=0.8\linewidth]{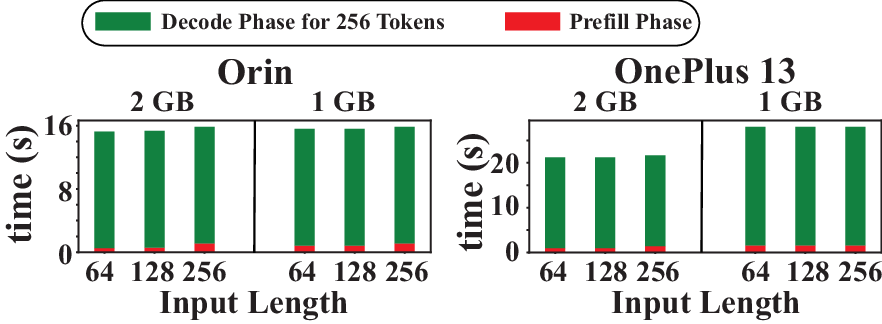}  
        \vspace{-0.3cm}
\caption{Impact of various input prefill lengths on total inference latency on the Qasper dataset using Mistral-7B on Jetson AGX Orin and OnePlus 13 with a fixed output of 256 tokens.}
        \label{fig:prefill}
\end{figure}

Figure~\ref{fig:prefill} illustrates the impact of input prefill length on the total inference latency of Mistral-7B on Jetson AGX Orin and OnePlus 13. The results show that prefill overhead is mainly determined by the initial dense weight loading and the corresponding computation. While prefill may involve a large number of input tokens, it has a substantially higher compute-to-I/O ratio than decoding. Even for a 256-token prompt followed by 256-token generation, decoding remains the dominant contributor to total inference latency. This suggests that, on edge devices, the sequential and memory-intensive decoding phase is a more critical bottleneck than prompt prefill.

\vspace{-0.2cm}
\subsubsection{Energy Consumption and Thermal Behavior}
\label{sec:energy_thermal}

Table~\ref{tab:energy_thermal} reports the energy consumption and thermal behavior during sustained inference. To eliminate variations caused by DVFS and ensure a fair comparison, we fix the CPU and GPU frequencies at 2201~MHz and 713~MHz, respectively, for all systems. LeanStream consumes slightly more energy per token than PowerInfer-2 and DejaVu because it maintains higher CPU and GPU utilization to overlap computation with weight loading. This also leads to higher peak CPU and GPU temperatures. LeanStream reaches a peak GPU temperature of $76^\circ$C, which remains below the $99^\circ$C thermal-throttling threshold. None of the evaluated systems experiences thermal throttling during the two-hour execution.

\vspace{-0.25cm}
\section{Conclusion}
We presented \emph{LeanStream}, a streaming speculate-and-refine framework for on-device LLM inference on resource-constrained mobile and embedded platforms. By progressively refining execution, loading, and cache decisions from partial GPU results, LeanStream resolves the tension between accurate context-aware prediction and efficient computation.

\begin{table}[!t]
\centering

\footnotesize
\setlength{\tabcolsep}{3.5pt}
\renewcommand{\arraystretch}{0.95}

\caption{Energy consumption and thermal behavior on Jetson AGX Orin. Mistral-7B on Scrolls-Qasper with 3~GB memory budget. Each system runs continuously for two hours at an ambient temperature of $21^\circ$C.}
\label{tab:energy_thermal}
\vspace{-0.2cm}
\begin{tabular}{lcccc}
\toprule
System &
\shortstack{Energy/Token \\ (J)} &
\shortstack{Peak GPU \\ Temp. ($^\circ$C)} &
\shortstack{Peak CPU \\ Temp. ($^\circ$C)} &
\shortstack{Thermal \\ Throttling} \\
\midrule
LeanStream   & 3.9 & 76 & 69 & No \\
PowerInfer-2 & 3.5 & 63 & 66 & No \\
DejaVu       & 3.7 & 60 & 65 & No \\
\bottomrule
\end{tabular}
\end{table}

\vspace{-0.25cm}
\section{Disclaimer}
\vspace{-0.1cm}
This paper was prepared for informational purposes with contributions from the Global Technology Applied Research center of JPMorgan Chase \& Co. (JPMC) and is not a product of its, or its affiliates’, Research Departments. JPMC and its affiliates make no representations or warranties, express or implied, regarding the completeness, accuracy, or reliability of the information herein, and accept no liability for its use or any related outcomes. This document does not constitute investment advice, financial research, or a recommendation or offer to
buy or sell any security, financial instrument, product, or service.
\vspace{-0.2cm}
\section{Acknowledgements}
This work is in part supported by the National Science Foundation grants III-2107200, CNS-2038658 and CNS-2038923.

}


\newpage

\bibliographystyle{unsrtnat}
\bibliography{references}

\begin{thebibliography}{68}
\providecommand{\natexlab}[1]{#1}
\providecommand{\url}[1]{\texttt{#1}}
\expandafter\ifx\csname urlstyle\endcsname\relax
  \providecommand{\doi}[1]{doi: #1}\else
  \providecommand{\doi}{doi: \begingroup \urlstyle{rm}\Url}\fi

\bibitem[Alizadeh et~al.(2024)Alizadeh, Mirzadeh, Belenko, Khatamifard, Cho, Del~Mundo, Rastegari, and Farajtabar]{llmflash}
Keivan Alizadeh, Seyed~Iman Mirzadeh, Dmitry Belenko, S~Khatamifard, Minsik Cho, Carlo~C Del~Mundo, Mohammad Rastegari, and Mehrdad Farajtabar.
\newblock Llm in a flash: Efficient large language model inference with limited memory.
\newblock In \emph{Proceedings of the 62nd Annual Meeting of the Association for Computational Linguistics (Volume 1: Long Papers)}, pages 12562--12584, 2024.

\bibitem[Song et~al.(2024)Song, Mi, Xie, and Chen]{powerinfer}
Yixin Song, Zeyu Mi, Haotong Xie, and Haibo Chen.
\newblock Powerinfer: Fast large language model serving with a consumer-grade gpu.
\newblock In \emph{Proceedings of the ACM SIGOPS 30th Symposium on Operating Systems Principles}, pages 590--606, 2024.

\bibitem[Xue et~al.(2024)Xue, Song, Mi, Zheng, Xia, and Chen]{powerinfer2}
Zhenliang Xue, Yixin Song, Zeyu Mi, Xinrui Zheng, Yubin Xia, and Haibo Chen.
\newblock Powerinfer-2: Fast large language model inference on a smartphone.
\newblock \emph{arXiv preprint arXiv:2406.06282}, 2024.

\bibitem[Liu et~al.(2023)Liu, Wang, Dao, Zhou, Yuan, Song, Shrivastava, Zhang, Tian, Re, et~al.]{dejavu}
Zichang Liu, Jue Wang, Tri Dao, Tianyi Zhou, Binhang Yuan, Zhao Song, Anshumali Shrivastava, Ce~Zhang, Yuandong Tian, Christopher Re, et~al.
\newblock Deja vu: Contextual sparsity for efficient llms at inference time.
\newblock In \emph{International Conference on Machine Learning}, pages 22137--22176. PMLR, 2023.

\bibitem[Liu et~al.(2025{\natexlab{a}})Liu, Ponnusamy, Cai, Guo, Kim, and Athiwaratkun]{liu2025trainingfree}
James Liu, Pragaash Ponnusamy, Tianle Cai, Han Guo, Yoon Kim, and Ben Athiwaratkun.
\newblock Training-free activation sparsity in large language models.
\newblock In \emph{The Thirteenth International Conference on Learning Representations}, 2025{\natexlab{a}}.
\newblock URL \url{https://openreview.net/forum?id=dGVZwyq5tV}.

\bibitem[Federici et~al.(2025)Federici, Belli, Van~Baalen, Jalalirad, Skliar, Major, Nagel, and Whatmough]{federici2025efficient}
Marco Federici, Davide Belli, Mart Van~Baalen, Amir Jalalirad, Andrii Skliar, Bence Major, Markus Nagel, and Paul Whatmough.
\newblock Efficient llm inference using dynamic input pruning and cache-aware masking.
\newblock \emph{Proceedings of Machine Learning and Systems}, 7, 2025.

\bibitem[Guo et~al.(2023)Guo, Choe, and Lin]{guo2023sti}
Liwei Guo, Wonkyo Choe, and Felix~Xiaozhu Lin.
\newblock Sti: Turbocharge nlp inference at the edge via elastic pipelining.
\newblock In \emph{Proceedings of the 28th ACM International Conference on Architectural Support for Programming Languages and Operating Systems, Volume 2}, pages 791--803, 2023.

\bibitem[Wang et~al.(2025)Wang, Zhou, Hong, and Guo]{wang2025d2moe}
Haodong Wang, Qihua Zhou, Zicong Hong, and Song Guo.
\newblock D2moe: Dual routing and dynamic scheduling for efficient on-device moe-based llm serving.
\newblock In \emph{Proceedings of the 31st Annual International Conference on Mobile Computing and Networking}, pages 574--588, 2025.

\bibitem[Chen et~al.(2025)Chen, Xie, Zhang, Tang, Wang, Dong, Chen, Yuan, Lin, Qiu, et~al.]{chen2025ktransformers}
Hongtao Chen, Weiyu Xie, Boxin Zhang, Jingqi Tang, Jiahao Wang, Jianwei Dong, Shaoyuan Chen, Ziwei Yuan, Chen Lin, Chengyu Qiu, et~al.
\newblock Ktransformers: Unleashing the full potential of cpu/gpu hybrid inference for moe models.
\newblock In \emph{Proceedings of the ACM SIGOPS 31st Symposium on Operating Systems Principles}, pages 1014--1029, 2025.

\bibitem[Chen et~al.(2026)Chen, Du, Liu, Yao, Yan, Liao, Liu, Wu, and Chen]{chen2026tokenflow}
Junyi Chen, Chuheng Du, Renyuan Liu, Shuochao Yao, Dingtian Yan, Jiang Liao, Shengzhong Liu, Fan Wu, and Guihai Chen.
\newblock Tokenflow: Responsive llm text streaming serving under request burst via preemptive scheduling.
\newblock In \emph{Proceedings of the 21st European Conference on Computer Systems}, pages 497--513, 2026.

\bibitem[Jiang et~al.(2023)Jiang, Li, Zhu, Li, Deng, Han, Shen, Shi, and Zhang]{mistral}
Yihang Jiang, Xiaoyang Li, Guangxu Zhu, Hang Li, Jing Deng, Kaifeng Han, Chao Shen, Qingjiang Shi, and Rui Zhang.
\newblock 6g non-terrestrial networks enabled low-altitude economy: Opportunities and challenges.
\newblock \emph{arXiv preprint arXiv:2311.09047}, 2023.

\bibitem[Touvron et~al.(2023)Touvron, Martin, Stone, Albert, Almahairi, Babaei, Bashlykov, Batra, Bhargava, Bhosale, et~al.]{llama}
Hugo Touvron, Louis Martin, Kevin Stone, Peter Albert, Amjad Almahairi, Yasmine Babaei, Nikolay Bashlykov, Soumya Batra, Prajjwal Bhargava, Shruti Bhosale, et~al.
\newblock Llama 2: Open foundation and fine-tuned chat models.
\newblock \emph{arXiv preprint arXiv:2307.09288}, 2023.

\bibitem[Yang et~al.(2025)Yang, Li, Yang, Zhang, Hui, Zheng, Yu, Gao, Huang, Lv, et~al.]{qwen}
An~Yang, Anfeng Li, Baosong Yang, Beichen Zhang, Binyuan Hui, Bo~Zheng, Bowen Yu, Chang Gao, Chengen Huang, Chenxu Lv, et~al.
\newblock Qwen3 technical report.
\newblock \emph{arXiv preprint arXiv:2505.09388}, 2025.

\bibitem[Yao et~al.(2017)Yao, Zhao, Zhang, Su, and Abdelzaher]{yao2017deepiot}
Shuochao Yao, Yiran Zhao, Aston Zhang, Lu~Su, and Tarek Abdelzaher.
\newblock Deepiot: Compressing deep neural network structures for sensing systems with a compressor-critic framework.
\newblock In \emph{Proceedings of the 15th ACM conference on embedded network sensor systems}, pages 1--14, 2017.

\bibitem[Yao et~al.(2018)Yao, Zhao, Shao, Liu, Liu, Su, and Abdelzaher]{yao2018fastdeepiot}
Shuochao Yao, Yiran Zhao, Huajie Shao, ShengZhong Liu, Dongxin Liu, Lu~Su, and Tarek Abdelzaher.
\newblock Fastdeepiot: Towards understanding and optimizing neural network execution time on mobile and embedded devices.
\newblock In \emph{Proceedings of the 16th ACM Conference on Embedded Networked Sensor Systems}, pages 278--291, 2018.

\bibitem[Liu et~al.(2024{\natexlab{a}})Liu, Leng, Tian, Hu, Chen, and Yao]{liu2024dynaspa}
Renyuan Liu, Yuyang Leng, Shilei Tian, Shaohan Hu, Chun-Fu Chen, and Shuochao Yao.
\newblock Dynaspa: Exploiting spatial sparsity for efficient dynamic dnn inference on devices.
\newblock In \emph{Proceedings of the 22nd ACM Conference on Embedded Networked Sensor Systems}, pages 422--435, 2024{\natexlab{a}}.

\bibitem[Liu et~al.(2025{\natexlab{b}})Liu, Leng, Liu, Hu, Chen, Zhao, Yun, and Yao]{liu2025daf}
Renyuan Liu, Yuyang Leng, Kaiyan Liu, Shaohan Hu, Chun-Fu Chen, Peijun Zhao, Heechul Yun, and Shuochao Yao.
\newblock Daf: An efficient end-to-end dynamic activation framework for on-device dnn training.
\newblock In \emph{Proceedings of the 23rd Annual International Conference on Mobile Systems, Applications and Services}, pages 196--208, 2025{\natexlab{b}}.

\bibitem[Liu et~al.(2025{\natexlab{c}})Liu, Leng, Tian, Hu, Chen, and Yao]{liu2025device}
Renyuan Liu, Yuyang Leng, Shilei Tian, Shaohan Hu, Richard Chen, and Shuochao Yao.
\newblock On-device dynamic dnn inference through spatial sparsity exploitation.
\newblock \emph{GetMobile: Mobile Computing and Communications}, 29\penalty0 (3):\penalty0 35--38, 2025{\natexlab{c}}.

\bibitem[Leng et~al.(2023)Leng, Liu, Guo, Chen, and Yao]{leng2023scaleflow}
Yuyang Leng, Renyuan Liu, Hongpeng Guo, Songqing Chen, and Shuochao Yao.
\newblock Scaleflow: Efficient deep vision pipeline with closed-loop scale-adaptive inference.
\newblock In \emph{Proceedings of the 31st ACM International Conference on Multimedia}, pages 1698--1706, 2023.

\bibitem[Leng et~al.(2026)Leng, Liu, Hu, Zhao, Chen, Chen, and Yao]{leng2026physical}
Yuyang Leng, Renyuan Liu, Shaohan Hu, Peijun Zhao, Chun-Fu Chen, Songqing Chen, and Shuochao Yao.
\newblock Physical self-supervised learning: Imu sensing without manual labels.
\newblock In \emph{Proceedings of the 24th Annual International Conference on Mobile Systems, Applications and Services}, pages 1011--1025, 2026.

\bibitem[Roumeliotis et~al.(2023)Roumeliotis, Tselikas, and Nasiopoulos]{roumeliotis2023llama}
Konstantinos~I Roumeliotis, Nikolaos~D Tselikas, and Dimitrios~K Nasiopoulos.
\newblock Llama 2: Early adopters' utilization of meta's new open-source pretrained model.
\newblock 2023.

\bibitem[Song et~al.(2025)Song, Han, Zhang, Hu, Shi, Li, Chen, Liu, Li, Yang, et~al.]{song2025prosparse}
Chenyang Song, Xu~Han, Zhengyan Zhang, Shengding Hu, Xiyu Shi, Kuai Li, Chen Chen, Zhiyuan Liu, Guangli Li, Tao Yang, et~al.
\newblock Prosparse: Introducing and enhancing intrinsic activation sparsity within large language models.
\newblock In \emph{Proceedings of the 31st International Conference on Computational Linguistics}, pages 2626--2644, 2025.

\bibitem[Zhang et~al.(2022)Zhang, Lin, Liu, Li, Sun, and Zhou]{zhang2022moefication}
Zhengyan Zhang, Yankai Lin, Zhiyuan Liu, Peng Li, Maosong Sun, and Jie Zhou.
\newblock Moefication: Transformer feed-forward layers are mixtures of experts.
\newblock In \emph{Findings of the Association for Computational Linguistics: ACL 2022}, pages 877--890, 2022.

\bibitem[Narayanan et~al.(2019)Narayanan, Harlap, Phanishayee, Seshadri, Devanur, Ganger, Gibbons, and Zaharia]{narayanan2019pipedream}
Deepak Narayanan, Aaron Harlap, Amar Phanishayee, Vivek Seshadri, Nikhil~R Devanur, Gregory~R Ganger, Phillip~B Gibbons, and Matei Zaharia.
\newblock Pipedream: Generalized pipeline parallelism for dnn training.
\newblock In \emph{Proceedings of the 27th ACM symposium on operating systems principles}, pages 1--15, 2019.

\bibitem[Wang et~al.(2022)Wang, Wei, Sabne, Davis, Ilbeyi, Hechtman, Chen, Murthy, Maggioni, Zhang, et~al.]{wang2022overlap}
Shibo Wang, Jinliang Wei, Amit Sabne, Andy Davis, Berkin Ilbeyi, Blake Hechtman, Dehao Chen, Karthik~Srinivasa Murthy, Marcello Maggioni, Qiao Zhang, et~al.
\newblock Overlap communication with dependent computation via decomposition in large deep learning models.
\newblock In \emph{Proceedings of the 28th ACM International Conference on Architectural Support for Programming Languages and Operating Systems, Volume 1}, pages 93--106, 2022.

\bibitem[Chen et~al.(2024)Chen, Li, Zhu, Duan, Sun, Zhang, and Yang]{chen2024centauri}
Chang Chen, Xiuhong Li, Qianchao Zhu, Jiangfei Duan, Peng Sun, Xingcheng Zhang, and Chao Yang.
\newblock Centauri: Enabling efficient scheduling for communication-computation overlap in large model training via communication partitioning.
\newblock In \emph{Proceedings of the 29th ACM International Conference on Architectural Support for Programming Languages and Operating Systems, Volume 3}, pages 178--191, 2024.

\bibitem[Bae et~al.(2021)Bae, Lee, Jin, Son, Kim, Jang, Ham, and Lee]{bae2021flashneuron}
Jonghyun Bae, Jongsung Lee, Yunho Jin, Sam Son, Shine Kim, Hakbeom Jang, Tae~Jun Ham, and Jae~W Lee.
\newblock $\{$FlashNeuron$\}$:$\{$SSD-Enabled$\}$$\{$Large-Batch$\}$ training of very deep neural networks.
\newblock In \emph{19th USENIX conference on file and storage technologies (FAST 21)}, pages 387--401, 2021.

\bibitem[Rajbhandari et~al.(2021)Rajbhandari, Ruwase, Rasley, Smith, and He]{rajbhandari2021zero}
Samyam Rajbhandari, Olatunji Ruwase, Jeff Rasley, Shaden Smith, and Yuxiong He.
\newblock Zero-infinity: Breaking the gpu memory wall for extreme scale deep learning.
\newblock In \emph{Proceedings of the international conference for high performance computing, networking, storage and analysis}, pages 1--14, 2021.

\bibitem[Jeong et~al.(2013)Jeong, Lee, Lee, Son, and Won]{jeong2013stack}
Sooman Jeong, Kisung Lee, Seongjin Lee, Seoungbum Son, and Youjip Won.
\newblock $\{$I/O$\}$ stack optimization for smartphones.
\newblock In \emph{2013 USENIX Annual Technical Conference (USENIX ATC 13)}, pages 309--320, 2013.

\bibitem[Agrawal et~al.(2008)Agrawal, Prabhakaran, Wobber, Davis, Manasse, and Panigrahy]{agrawal2008design}
Nitin Agrawal, Vijayan Prabhakaran, Ted Wobber, John~D Davis, Mark Manasse, and Rina Panigrahy.
\newblock Design tradeoffs for $\{$SSD$\}$ performance.
\newblock In \emph{2008 USENIX Annual Technical Conference (USENIX ATC 08)}, 2008.

\bibitem[Ji et~al.(2016)Ji, Chang, Shi, Wu, Li, and Xue]{ji2016empirical}
Cheng Ji, Li-Pin Chang, Liang Shi, Chao Wu, Qiao Li, and Chun~Jason Xue.
\newblock An empirical study of $\{$File-System$\}$ fragmentation in mobile storage systems.
\newblock In \emph{8th USENIX Workshop on Hot Topics in Storage and File Systems (HotStorage 16)}, 2016.

\bibitem[Mikolov et~al.(2013)Mikolov, Sutskever, Chen, Corrado, and Dean]{mikolov2013distributed}
Tomas Mikolov, Ilya Sutskever, Kai Chen, Greg~S Corrado, and Jeff Dean.
\newblock Distributed representations of words and phrases and their compositionality.
\newblock \emph{Advances in neural information processing systems}, 26, 2013.

\bibitem[Pennington et~al.(2014)Pennington, Socher, and Manning]{pennington2014glove}
Jeffrey Pennington, Richard Socher, and Christopher~D Manning.
\newblock Glove: Global vectors for word representation.
\newblock In \emph{Proceedings of the 2014 conference on empirical methods in natural language processing (EMNLP)}, pages 1532--1543, 2014.

\bibitem[Salakhutdinov and Hinton(2007)]{salakhutdinov2007learning}
Ruslan Salakhutdinov and Geoff Hinton.
\newblock Learning a nonlinear embedding by preserving class neighbourhood structure.
\newblock In \emph{Artificial intelligence and statistics}, pages 412--419. PMLR, 2007.

\bibitem[Kjolstad et~al.(2017)Kjolstad, Kamil, Chou, Lugato, and Amarasinghe]{taco}
Fredrik Kjolstad, Shoaib Kamil, Stephen Chou, David Lugato, and Saman Amarasinghe.
\newblock The tensor algebra compiler.
\newblock \emph{Proceedings of the ACM on Programming Languages}, 1\penalty0 (OOPSLA):\penalty0 1--29, 2017.

\bibitem[Liu et~al.(2024{\natexlab{b}})Liu, Leng, Tian, Hu, Chen, and Yao]{dynaspa}
Renyuan Liu, Yuyang Leng, Shilei Tian, Shaohan Hu, Chun-Fu Chen, and Shuochao Yao.
\newblock Dynaspa: Exploiting spatial sparsity for efficient dynamic dnn inference on devices.
\newblock In \emph{Proceedings of the 22nd ACM Conference on Embedded Networked Sensor Systems}, pages 422--435, 2024{\natexlab{b}}.

\bibitem[Mayne et~al.(2000)Mayne, Rawlings, Rao, and Scokaert]{mayne2000constrained}
David~Q Mayne, James~B Rawlings, Christopher~V Rao, and Pierre~OM Scokaert.
\newblock Constrained model predictive control: Stability and optimality.
\newblock \emph{Automatica}, 36\penalty0 (6):\penalty0 789--814, 2000.

\bibitem[Mesbah(2016)]{mesbah2016stochastic}
Ali Mesbah.
\newblock Stochastic model predictive control: An overview and perspectives for future research.
\newblock \emph{IEEE Control Systems Magazine}, 36\penalty0 (6):\penalty0 30--44, 2016.

\bibitem[Bemporad et~al.(2002)Bemporad, Morari, Dua, and Pistikopoulos]{bemporad2002explicit}
Alberto Bemporad, Manfred Morari, Vivek Dua, and Efstratios~N Pistikopoulos.
\newblock The explicit linear quadratic regulator for constrained systems.
\newblock \emph{Automatica}, 38\penalty0 (1):\penalty0 3--20, 2002.

\bibitem[Marescot et~al.(2013)Marescot, Chapron, Chad{\`e}s, Fackler, Duchamp, Marboutin, and Gimenez]{marescot2013complex}
Lucile Marescot, Guillaume Chapron, Iadine Chad{\`e}s, Paul~L Fackler, Christophe Duchamp, Eric Marboutin, and Olivier Gimenez.
\newblock Complex decisions made simple: a primer on stochastic dynamic programming.
\newblock \emph{Methods in Ecology and Evolution}, 4\penalty0 (9):\penalty0 872--884, 2013.

\bibitem[Indyk and Motwani(1998)]{indyk1998approximate}
Piotr Indyk and Rajeev Motwani.
\newblock Approximate nearest neighbors: towards removing the curse of dimensionality.
\newblock In \emph{Proceedings of the thirtieth annual ACM symposium on Theory of computing}, pages 604--613, 1998.

\bibitem[Gionis et~al.(1999)Gionis, Indyk, Motwani, et~al.]{gionis1999similarity}
Aristides Gionis, Piotr Indyk, Rajeev Motwani, et~al.
\newblock Similarity search in high dimensions via hashing.
\newblock In \emph{Vldb}, volume~99, pages 518--529, 1999.

\bibitem[Yagnik et~al.(2011)Yagnik, Strelow, Ross, and Lin]{yagnik2011power}
Jay Yagnik, Dennis Strelow, David~A Ross, and Ruei-sung Lin.
\newblock The power of comparative reasoning.
\newblock In \emph{2011 International Conference on Computer Vision}, pages 2431--2438. IEEE, 2011.

\bibitem[Zeng et~al.(2023)Zeng, Davies, Pulijala, Sankaralingam, and Singh]{zeng2023lookupffn}
Zhanpeng Zeng, Michael Davies, Pranav Pulijala, Karthikeyan Sankaralingam, and Vikas Singh.
\newblock Lookupffn: making transformers compute-lite for cpu inference.
\newblock In \emph{International Conference on Machine Learning}, pages 40707--40718. PMLR, 2023.

\bibitem[Blalock and Guttag(2021)]{blalock2021multiplying}
Davis Blalock and John Guttag.
\newblock Multiplying matrices without multiplying.
\newblock In \emph{International Conference on Machine Learning}, pages 992--1004. PMLR, 2021.

\bibitem[Wei et~al.(2025)Wei, Cao, Cao, Ma, Wang, Zhang, and Yang]{wei2025t}
Jianyu Wei, Shijie Cao, Ting Cao, Lingxiao Ma, Lei Wang, Yanyong Zhang, and Mao Yang.
\newblock T-mac: Cpu renaissance via table lookup for low-bit llm deployment on edge.
\newblock In \emph{Proceedings of the Twentieth European Conference on Computer Systems}, pages 278--292, 2025.

\bibitem[Tang et~al.(2023)Tang, Wang, Cao, Zhang, Chen, Cai, Liu, and Yang]{tang2023lut}
Xiaohu Tang, Yang Wang, Ting Cao, Li~Lyna Zhang, Qi~Chen, Deng Cai, Yunxin Liu, and Mao Yang.
\newblock Lut-nn: Empower efficient neural network inference with centroid learning and table lookup.
\newblock In \emph{Proceedings of the 29th Annual International Conference on Mobile Computing and Networking}, pages 1--15, 2023.

\bibitem[Spring and Shrivastava(2017)]{spring2017scalable}
Ryan Spring and Anshumali Shrivastava.
\newblock Scalable and sustainable deep learning via randomized hashing.
\newblock In \emph{Proceedings of the 23rd ACM SIGKDD International Conference on Knowledge Discovery and Data Mining}, pages 445--454, 2017.

\bibitem[Chen et~al.(2020{\natexlab{a}})Chen, Medini, Farwell, Tai, Shrivastava, et~al.]{chen2020slide}
Beidi Chen, Tharun Medini, James Farwell, Charlie Tai, Anshumali Shrivastava, et~al.
\newblock Slide: In defense of smart algorithms over hardware acceleration for large-scale deep learning systems.
\newblock \emph{Proceedings of Machine Learning and Systems}, 2:\penalty0 291--306, 2020{\natexlab{a}}.

\bibitem[Chen et~al.(2020{\natexlab{b}})Chen, Liu, Peng, Xu, Li, Dao, Song, Shrivastava, and Re]{chen2020mongoose}
Beidi Chen, Zichang Liu, Binghui Peng, Zhaozhuo Xu, Jonathan~Lingjie Li, Tri Dao, Zhao Song, Anshumali Shrivastava, and Christopher Re.
\newblock Mongoose: A learnable lsh framework for efficient neural network training.
\newblock In \emph{International Conference on Learning Representations}, 2020{\natexlab{b}}.

\bibitem[Andoni and Razenshteyn(2015)]{andoni2015optimal}
Alexandr Andoni and Ilya Razenshteyn.
\newblock Optimal data-dependent hashing for approximate near neighbors.
\newblock In \emph{Proceedings of the forty-seventh annual ACM symposium on Theory of computing}, pages 793--801, 2015.

\bibitem[Dong et~al.(2019)Dong, Indyk, Razenshteyn, and Wagner]{dong2019learning}
Yihe Dong, Piotr Indyk, Ilya Razenshteyn, and Tal Wagner.
\newblock Learning space partitions for nearest neighbor search.
\newblock \emph{arXiv preprint arXiv:1901.08544}, 2019.

\bibitem[Andoni et~al.(2015)Andoni, Indyk, Laarhoven, Razenshteyn, and Schmidt]{andoni2015practical}
Alexandr Andoni, Piotr Indyk, Thijs Laarhoven, Ilya Razenshteyn, and Ludwig Schmidt.
\newblock Practical and optimal lsh for angular distance.
\newblock \emph{Advances in neural information processing systems}, 28, 2015.

\bibitem[Blalock and Guttag(2017)]{blalock2017bolt}
Davis~W Blalock and John~V Guttag.
\newblock Bolt: Accelerated data mining with fast vector compression.
\newblock In \emph{Proceedings of the 23rd ACM SIGKDD International Conference on Knowledge Discovery and Data Mining}, pages 727--735, 2017.

\bibitem[Wang et~al.(2017)Wang, Zhang, Sebe, Shen, et~al.]{wang2017survey}
Jingdong Wang, Ting Zhang, Nicu Sebe, Heng~Tao Shen, et~al.
\newblock A survey on learning to hash.
\newblock \emph{IEEE transactions on pattern analysis and machine intelligence}, 40\penalty0 (4):\penalty0 769--790, 2017.

\bibitem[Hubara et~al.(2016)Hubara, Courbariaux, Soudry, El-Yaniv, and Bengio]{hubara2016binarized}
Itay Hubara, Matthieu Courbariaux, Daniel Soudry, Ran El-Yaniv, and Yoshua Bengio.
\newblock Binarized neural networks.
\newblock \emph{Advances in neural information processing systems}, 29, 2016.

\bibitem[Xiao et~al.(2023)Xiao, Lin, Seznec, Wu, Demouth, and Han]{xiao2023smoothquant}
Guangxuan Xiao, Ji~Lin, Mickael Seznec, Hao Wu, Julien Demouth, and Song Han.
\newblock Smoothquant: Accurate and efficient post-training quantization for large language models.
\newblock In \emph{International conference on machine learning}, pages 38087--38099. PMLR, 2023.

\bibitem[Charikar(2002)]{charikar2002similarity}
Moses~S Charikar.
\newblock Similarity estimation techniques from rounding algorithms.
\newblock In \emph{Proceedings of the thiry-fourth annual ACM symposium on Theory of computing}, pages 380--388, 2002.

\bibitem[Gray and Neuhoff(2002)]{gray2002quantization}
Robert~M. Gray and David~L. Neuhoff.
\newblock Quantization.
\newblock \emph{IEEE transactions on information theory}, 44\penalty0 (6):\penalty0 2325--2383, 2002.

\bibitem[Bengio et~al.(2013)Bengio, L{\'e}onard, and Courville]{bengio2013estimating}
Yoshua Bengio, Nicholas L{\'e}onard, and Aaron Courville.
\newblock Estimating or propagating gradients through stochastic neurons for conditional computation.
\newblock \emph{arXiv preprint arXiv:1308.3432}, 2013.

\bibitem[Jang et~al.(2016)Jang, Gu, and Poole]{jang2016categorical}
Eric Jang, Shixiang Gu, and Ben Poole.
\newblock Categorical reparameterization with gumbel-softmax.
\newblock \emph{arXiv preprint arXiv:1611.01144}, 2016.

\bibitem[Abdi and Williams(2010)]{abdi2010principal}
Herv{\'e} Abdi and Lynne~J Williams.
\newblock Principal component analysis.
\newblock \emph{Wiley interdisciplinary reviews: computational statistics}, 2\penalty0 (4):\penalty0 433--459, 2010.

\bibitem[Lee et~al.(2024)Lee, Lee, Seo, and Sim]{lee2024infinigen}
Wonbeom Lee, Jungi Lee, Junghwan Seo, and Jaewoong Sim.
\newblock $\{$InfiniGen$\}$: Efficient generative inference of large language models with dynamic $\{$KV$\}$ cache management.
\newblock In \emph{18th USENIX Symposium on Operating Systems Design and Implementation (OSDI 24)}, pages 155--172, 2024.

\bibitem[Song et~al.(2020)Song, Berger, Li, and Lloyd]{song2020learning}
Zhenyu Song, Daniel~S Berger, Kai Li, and Wyatt Lloyd.
\newblock Learning relaxed belady for content distribution network caching.
\newblock In \emph{17th USENIX Symposium on Networked Systems Design and Implementation (NSDI 20)}, pages 529--544, 2020.

\bibitem[Yang et~al.(2023)Yang, Mao, Yue, and Rashmi]{yang2023gl}
Juncheng Yang, Ziming Mao, Yao Yue, and KV~Rashmi.
\newblock $\{$GL-Cache$\}$: Group-level learning for efficient and high-performance caching.
\newblock In \emph{21st USENIX Conference on File and Storage Technologies (FAST 23)}, pages 115--134, 2023.

\bibitem[Shaham et~al.(2022)Shaham, Segal, Ivgi, Efrat, Yoran, Haviv, Gupta, Xiong, Geva, Berant, et~al.]{scrolls}
Uri Shaham, Elad Segal, Maor Ivgi, Avia Efrat, Ori Yoran, Adi Haviv, Ankit Gupta, Wenhan Xiong, Mor Geva, Jonathan Berant, et~al.
\newblock Scrolls: Standardized comparison over long language sequences.
\newblock In \emph{Proceedings of the 2022 Conference on Empirical Methods in Natural Language Processing}, pages 12007--12021, 2022.

\bibitem[Lin et~al.(2022)Lin, Hilton, and Evans]{truthfulqa}
Stephanie Lin, Jacob Hilton, and Owain Evans.
\newblock Truthfulqa: Measuring how models mimic human falsehoods.
\newblock In \emph{Proceedings of the 60th annual meeting of the association for computational linguistics (volume 1: long papers)}, pages 3214--3252, 2022.

\bibitem[Reddy et~al.(2019)Reddy, Chen, and Manning]{coqa}
Siva Reddy, Danqi Chen, and Christopher~D Manning.
\newblock Coqa: A conversational question answering challenge.
\newblock \emph{Transactions of the Association for Computational Linguistics}, 7:\penalty0 249--266, 2019.

\end{thebibliography}
\newpage
\clearpage

\end{document}